\documentclass[aps,pre,onecolumn,superscriptaddress,amssymb]{revtex4}

\usepackage[T1]{fontenc}
\usepackage{graphicx}

\usepackage{makeidx}
\usepackage{amsmath}
\usepackage{wrapfig}
\usepackage{comment}
\usepackage{hyperref}
\usepackage[dvipsnames]{color}
\usepackage{braket}
\usepackage{amssymb}
\usepackage{bm}
\usepackage{xcolor}
\usepackage{txfonts}

\newcommand \be{\begin{eqnarray}}
\newcommand \ee{\end{eqnarray}}
\newcommand{\del}{\partial}
\newcommand{\idm}{\mathbf{1}}

\newcommand{\tr}{\textup{Tr}}
\newcommand{\cG}{{\cal G}}

\newcommand{\cX}{{\cal X}}

\newcommand{\cZ}{{\cal Z}}

\newcommand{\cR}{{\cal R}}

\newcommand{\cD}{{\cal D}}
\newcommand{\btr}{\textup{bTr}}

\newcommand{\<}{\left<}
\renewcommand{\>}{\right>}
\def\simge{\mathrel{
    \rlap{\raise 0.511ex \hbox{$>$}}{\lower 0.511ex \hbox{$\sim$}}}}
\def\simle{\mathrel{
    \rlap{\raise 0.511ex \hbox{$<$}}{\lower 0.511ex \hbox{$\sim$}}}}

\begin{document}

%

%
\title{Dynamical Isometry is Achieved in Residual Networks in a Universal Way for any Activation Function}

\author{Wojciech Tarnowski} 
\email{wojciech.tarnowski@uj.edu.pl} 
\affiliation{M. Smoluchowski Institute of Physics,  Jagiellonian University, PL--30--348 Krak\'ow, Poland}

\author{Piotr Warcho\l{}} 
\email{piotr.warchol@uj.edu.pl} 
\affiliation{M. Smoluchowski Institute of Physics,  Jagiellonian University, PL--30--348 Krak\'ow, Poland}

\author{Stanis\l aw Jastrz\k{e}bski} 
\email{stanislaw.jastrzebski@uj.edu.pl} 
\affiliation{Faculty of Mathematics and Computer Science, Jagiellonian University, Krak\'ow, Poland
}

\author{Jacek Tabor} 
\email{jcktbr@gmail.com} 
\affiliation{Faculty of Mathematics and Computer Science, Jagiellonian University, Krak\'ow, Poland
}

\author{Maciej A. Nowak} 
\email{maciej.a.nowak@uj.edu.pl} 
\affiliation{M. Smoluchowski Institute of Physics and Mark Kac Complex Systems Research Center,  Jagiellonian University, PL--30--348 Krak\'ow, Poland}

\date{\today}

\begin{abstract}
We demonstrate that in residual neural networks (ResNets) dynamical isometry is achievable irrespective of the activation function used. We do that by deriving, with the help of Free Probability and Random Matrix Theories, a universal formula for the spectral density of the input-output Jacobian at initialization, in the large network width and depth limit. The resulting singular value spectrum depends on a single parameter, which we calculate for a variety of popular activation functions, by analyzing the signal propagation in the artificial neural network. We corroborate our results with numerical simulations of both random matrices and ResNets applied to the CIFAR-10 classification problem. Moreover, we study consequences of this universal behavior for the initial and late phases of the learning processes. 
We conclude by drawing attention to the simple fact, that initialization acts as a confounding factor between the choice of activation function and the rate of learning. We propose that in ResNets this can be resolved based on our results by ensuring the same level of dynamical isometry at initialization.
\end{abstract}

\maketitle

\section{Introduction}
Deep Learning has achieved unparalleled success in fields such as object detection and recognition, language translation, and speech recognition \citep{LeCBH}. At the same time, models achieving these state-of-the-art results are increasingly deep and complex \citep{CCP}, which often leads to optimization challenges such as vanishing gradients. Many solutions to this problem have been proposed. In particular, Residual Neural Networks remedy this to some extent \citep{HZRS, VWB} by using skip connections in the network architecture, which improve gradient flow. As a result, Residual Neural Networks outmatched other competing models in the 2015 ILSVRC and COCO competitions. Yet another approach towards solving this problem is to tailor fit the networks weight initialization to facilitate training, for example by ensuring dynamical isometry \citep{PSG1}.  In this latter case, the insights are based on an analysis of the statistical properties of information propagation in the network and a study of the full singular spectrum of a particular matrix, namely the input-output Jacobian, via the techniques of Free Probability and Random Matrix Theories (FPT \& RMT). This perspective has recently led to successfully training a 10000 layer vanilla convolutional neural network \citep{XBSSP}.

RMT is a versatile tool that, since its inception, saw a substantial share of applications, from the earliest in nuclear physics \citep{Wigner} to the latest in game theory \citep{DysonGame} (see \citep{Oxford} for some of the use cases discovered in the mean time). It is thus not surprising that it found its way to be used to understand artificial neural networks. In particular, to study their loss surface \citep{CHMAL, PB}, the associated Gram matrix  \citep{LLC, PW} and in the case of single layer networks, their dynamics \citep{LC}. Our main contribution is extending the theoretical analysis of \citep{PSG1, SGGS, PSG2} to residual networks. In particular, we find that residual networks can achieve dynamical isometry for many different activation functions provided that the variance of weight initialization scale is inversely proportional to the number of skip-connections. This is in contrast to feedforward networks, where orthogonal weights and antisymmetric sigmoidal activation functions (like tanh) are required. These theoretical results are supported by an empirical investigation on the popular CIFAR-10 benchmark.

\subsection{Related work}

The framework of dynamical mean field theory, we will apply to study signal propagation in neural networks, was first used in this context in \citep{PLRSG}. There, the authors showed the existence of an order-to-chaos expressivity transition for deep feedforward neural networks with random initial weights, on the plane spanned by the variances of the network weights and biases. This in turn led to the insight of \citep{SGGS}, that arbitrary deep networks can be trained as long as they are close to the criticality associated with that transition. The techniques developed in these works, together with methods of FPT and RMT, allowed, for the first time, to analytically compute the singular value distribution of the input-output Jacobian of a deep feedforward network with nonlinear activation function and at criticality \citep{PSG1}. Finally, \citep{PSG2} showed, that for feedforward neural networks, in their large depth limit and at a special point of the above mentioned critical line, the singular spectrum of the Jacobian is given by a universal distribution depending on the form of the activation function used. In particular they distinguish the {\it Bernoulli}  and the {\it smooth} universality classes corresponding to piecewise linear and some nonlinear activation functions. In fact, in this paper, we take the approach of that last work and apply it to fully connected residual neural networks. We find a single universality class for this architecture.

Let us also mention some recent, important developments in the area of residual neural network initialization. One of the earlier developments, is the introduction of layer-sequential unit-variance (LSUV) initialization \citep{MM}. The two step process involved normalizing the outputs of the neurons on the first forward run and showed promising results. In another, very relevant paper \citep{Taki1}, analyzing the signal propagation in a similar manner to that mentioned in the paragraph above, shows for ResNets, with piecewise linear, symmetric as well as ReLU activation functions, that the proper variance for network weight initialization is of order $\frac{1}{NL}$, where $L$ is the number of layers and $N$ the number of neurons in each layer. A similar conclusion is reached by \citep{BFLLMM}. We corroborate this result with our analysis. Another contribution shows that adding skip connections to the network, eliminates the critical behavior described above \citep{YS1}. Finally, the importance of initialization in ResNets is shown by~\citep{OR}, where it is demonstrated that initializing to a zero function enables training state of the art residual networks without the use of batch normalization~\citep{IS}. Note that  ResNet with this initialization achieves in fact an ideal isometry.

When finishing this manuscript, we have learned of a recent paper tackling the same problem of ResNets initialization by studying the singular spectrum properties of the Jacobian with the tools of Free Probability. 
While the analysis of \citep{LQ} is related, it is crucial to note that the authors do not observe the universal character of the singular spectrum - the main result of our paper, and treat only piecewise linear activation functions. It is also worth mentioning that, similar to us, they rediscover the importance of $\frac{1}{LN}$ scaling of \citep{BFLLMM} and  \citep{Taki1}.

\subsection{Our results}
Our contributions are the following. We show that the singular spectrum of the input-output Jacobian, in the the networks large width and depth limit, is given by a universal formula - with the dependence on the type of activation function encapsulated in a single parameter.  Furthermore, we calculate the layer dependent statistical properties of the pre-activations for a variety of activation functions.  All together, this gives the associated singular spectra of the Jacobian, which we compare with random matrix and artificial neural network numerical simulations corroborating our theoretical results. The singular values of the input-output Jacobian concentrate around 1 for a wide range of parameters, which shows that fine-tuning the initialization is not required for achieving dynamical isometry in ResNets. Even though the final results of the theoretical calculations are derived in the limit $L,N\to\infty$, the numerical experiments match them already for  $L=10$ (with $N=500$). As a practical application of our work and the universality property it uncovers, we propose a framework for setting up weight initialization in experiments with residual neural networks.

\subsection{Structure of the paper}

We follow this introductory section by defining the model of ResNets we will work with and with a short note on the relevance of the input-output Jacobian. Then, in subsection \ref{s:spectrala}, we derive the equation governing the Green's function and hence the spectrum of the Jacobian, which depends on a single parameter, which we denote by $c$. Proceeding is the analysis of the propagation of the information in the network via an analysis of the probability density function describing the pre-activations across the layers at network initialization. This allows us to calculate $c$ for many different activation functions in Appendix \ref{s:actfunc}. We close the second section of the paper by revealing the random matrix experiments confirming our results. Sec.~\ref{s:Experiments}  is devoted to the outcome of associated residual neural network numerical calculations. There, we showcase the resulting, experimental, universal spectrum of the Jacobian and the outcomes of the learning processes. This is followed by a short comment, in Sec.~\ref{s:bn}, on the influence of batch normalization on the presented setup. We close the paper with a short discussion section. Finally, in the rest of the Appendices, we show the results of numerical experiments validating the signal propagation recurrence relations and some baseline (based on using the same weight matrix variances, irrespective of the choice of activation functions) simulations of the learning process.

\section{The model}
In this paper, we consider a deep, residual network of $L$ layers of a constant width of $N$ neurons. We follow the typical nomenclature of the literature and therefore the real-valued, synaptic matrix for the $l$-th layer is denoted by $\bm{W}^l$, whereas the real-valued bias vectors are $\bm{b}^l$. The information propagates in this network according to:

\be  \label{model}
\bm{x}^l=\phi(\bm{h}^l)+a \bm{x}^{l-1}, \,\,\,\, \bm{h}^l=\bm{W}^l \bm{x}^{l-1}+\bm{b}^l,
\ee
where $\bm{h}^l$ and $\bm{x}^l$ are pre- and post-activations respectively and $\phi$ is the activation function itself, acting entry-wise on the vector of pre-activations. We have introduced the parameter $a$ to track the influence of skip connections in the calculations, however we do not study its influence on the Jacobian's spectrum or learning in general. By $\bm{x}^0$ we denote the input of the network and by $\bm{x}^L$ its output.  
Our primary interest will lay in exploring the singular value spectral properties of the input-output Jacobian:
\be \label{jacobian}
J_{ik}=\frac{\partial x^L_i}{\partial x^0_k},
\ee
known to be useful in studying initialization schemes of neural networks at least since the work of \citep{GB}.   It in particular holds the information on the severity of the exploding gradients problem.

\subsection{Relevance of the input-output Jacobian}

To understand why we are interested in the Jacobian, consider the neural network adjusting its weights during the learning process. In a simplified, example by example scenario, this happens according to
\be 
\Delta W^l_{ij} = - \eta \frac{\del E(\bm{x}^L, \bm{y})}{\del W^l_{ij}},
\ee
where $E(\bm{x}^L,\bm{y})$ is the error function depending on $\bm{x}^L$ - the output of the network, $\bm{y}$ - the correct output value associated with that example and, implicitly through $\bm{x}^L$, on the parameters of the model, namely the weights and biases. Here, for simplicity, we consider only the adjustments of the weights - an analogous reasoning applies to the biases.  $\eta$ is the learning rate. By use of the chain rule we can rewrite this as:
\be 
\Delta W^l_{ij} = - \eta \sum_{k,t} \frac{\del x^l_t}{\del W^l_{ij}}   \frac{\del x^L_k}{\del x^l_t}  \frac{\del E(\bm{x}^L, \bm{y})}{\del x^L_{k}},
\ee 
For the learning process to be stable, all three terms need to be bounded. Out of those, the middle one can become problematic if a poor choice of the initialization scheme is made. We can rewrite it as:
\be \label{Jacobian1}
\frac{\del x^{L}_k}{\del x^{l}_t} = \left[\prod_{i=l+1}^{L} \left(\bm{D}^i \bm{W}^i+{\bm 1}a\right)  \right]_{kt}
\ee
and see the larger the difference between $L$ and $l$, the more terms we have in the product, and (in general) the less control there is over its behavior. Here ${\bm 1}$ is an identity matrix and, $\bm{D}^l$ is a diagonal matrix such that $D^l_{ij}=\phi'(h_i^l)\delta_{ij}$.
Indeed, it was proposed by \citep{GB}, that learning in deep feed-forward neural networks can be improved by keeping the mean singular value of the Jacobian associated with layer $i$ (in our setup $\bm{J}^i = \bm{D}^i \bm{W}^i+{\bm 1}a$), close to $1$ for all $i$'s.
It is also important for the dynamics of learning to be driven by data, not by the random initialization of the network. The latter may take place if the Jacobian to the $l$-th layer possesses very large singular values which dominate the learning or very small singular values suppressing it. In the optimal case all singular values should be concentrated around 1 regardless of how deep is the considered layer. One therefore examines the case of $l=0$, namely ${\del x^{L}_k}/{\del x^{0}_t}$ - the input-output Jacobian, as the most extreme object of (\ref{Jacobian1}). The feature that in the limit of large depth all singular values of $\bm{J}$ concentrate around 1, irrespective of the depth of the network, was coined as dynamical isometry~\citep{Saxe}.

Note, that the spectral problem for the full Jacobian 
\be \label{eq:jac1}
\bm{J}=\prod_{l=1}^{L}\left(\bm{D}^l\bm{W}^l+{\bm 1}a\right)
\ee
belongs to the class of  matrix-valued diffusion processes~\citep{GJJN,JW}, leading to a complex \textit{eigenvalue} spectrum. 
We note that the large $N$ limit, spectral properties of (\ref{eq:jac1}) with $\bm{D}={\bm 1}$ (deep linear networks), and different symmetry classes of ${\bm W}$, was derived already by \citep{GJJN}. Due to non-normality  of the Jacobian, singular values cannot be easily related to eigenvalues. Therefore we follow \citep{PSG1,PSG2} and tackle the full {\it singular spectrum} of  the Jacobian (or equivalently the eigenvalue spectrum of $\bm{JJ}^T$), extending these works to the case of the Residual Neural Network model.

\section{Spectral properties of the Jacobian}
\label{spectral}
\subsection{Spectral analysis}\label{s:spectrala}

Free Probability Theory, or Free Random Variable (FRV) Theory~\citep{VOICULESCU}, is a powerful tool for the spectral analysis of random matrices in the limit of their large size. It is a counterpart of the classical Probability Theory for the case of non-commuting observables. 
For a pedagogical introduction to the subject, see \citep{SpeicherMingo} - here we start by laying out the basics useful in the derivations of this subsection.
The fundamental objects of the theory are the Green's functions (a.k.a. Stieltjes transforms in mathematical literature):
\begin{equation}
G_H(z)=\<\frac{1}{N}\tr \left(z\idm-\bm{H}\right)^{-1}\> =\int_{-\infty}^{\infty}\frac{\rho_H(\lambda) d\lambda}{z-\lambda},
\end{equation}
which generate spectral moments and where the subscript $H$ indicates, that his formulation is proper for self-adjoint matrices. The eigenvalue density can be recovered via the Sochocki-Plemelj formula
\begin{equation}
\rho_H(x)=-\frac{1}{\pi}\lim_{\epsilon\to 0}G_H(x+i\epsilon).
\end{equation}
The associated free cumulants are generated  by the so-called $R$-transform, which plays the role of the logarithm of the characteristic function in the classical probability. 
By this correspondence, the $R$-transform  is  additive under addition, i.e. $R_{X+Y}(z)=R_{X}(z)+R_Y(z)$ for mutually free, but non-commuting random ensembles $X$ and $Y$. 
Moreover, it is related to $G$ via the functional equations
\begin{equation}
G\left(R(z)+\frac{1}{z}\right)=z,\qquad R(G(z))+\frac{1}{G(z)}=z. \label{eq:RGrelation}
\end{equation}
On the other hand, the so-called $S$-transform facilitates calculations of the spectra of products of random matrices, as it satisfies $S_{AB}(z)=S_A(z)S_B(z)$, provided $A$ and $B$ are mutually free and at least one is positive definite.  If additionally, the ensemble has a  finite mean, the S-transform  can be easily obtained from the $R$-transform, and vice versa, through a pair of the following, mutually inverse maps 
$z=yS(y)$ {\it and} $y=zR(z)$. Explicitly:  
\begin{equation}
S(zR(z))=\frac{1}{R(z)},\quad R(zS(z))=\frac{1}{S(z)}. \label{eq:RSrelation}
\end{equation}

Denoting now $\bm{J}_L$ the Jacobian across $L$ layers and $\bm{Y}_l=(a\mathbf{1}+\bm{D}^{l}\bm{W}^{l})$, one can write the recursion relation $\bm{J}_L\bm{J}_L^{T}=\bm{Y}_L\bm{J}_{L-1}\bm{J}_{L-1}^T\bm{Y}_L$. The latter matrix is isospectral to $\bm{Y}_{L}^T\bm{Y}_L \bm{J}_{L-1}\bm{J}_{L-1}^T$, which leads to the equation for the $S$-transform $S_{J_L J_L^T}(z)=S_{Y_L^T Y_L}(z)S_{J_{L-1}J_{L-1}^{T}}(z)$. Proceeding inductively, we arrive at

\begin{equation}
S_{JJ^T}(z)=\prod_{l=1}^{L}S_{Y_lY_l^T}(z). \label{eq:SProduct}
\end{equation}
To find the $S$-transform of  the single layer Jacobian, we will first consider its Green's function
\begin{equation}
G(z)=\<\frac{1}{N}\tr (z\idm-\bm{Y}_l\bm{Y}_l^T)^{-1}\>,
\end{equation}
with the averaging over the ensemble of weight matrices $\bm{W}^l$. To facilitate the study of $G$, in particular to cope with $\bm{YY}^T$, one linearizes the problem by introducing matrices of size $2N\times 2N$
\begin{equation}
\cZ:=\left(\begin{array}{cc}
-a & 1 \\
z & -a
\end{array}\right), \qquad \cX:=\left(\begin{array}{cc}
\bm{X} & 0 \\
0 & \bm{X}^T
\end{array}\right),
\end{equation}
with $\bm{X}=\bm{D}^l\bm{W}^l$. Another crucial ingredient is the block trace operation ($\btr$), which is the trace applied to each $N\times N$ block. The generalized Green's function is defined as a block trace of the generalized resolvent $(\cZ\otimes\idm-\cX)^{-1}$
\begin{equation}
\cG:=\left(\begin{array}{cc}
G_{11} & G_{12} \\
G_{21} & G_{22}
\end{array}\right) =
\<\frac{1}{N}\btr \left(\begin{array}{cc}
-a-\bm{X} & 1 \\
z & -a-\bm{X}^T
\end{array}\right)^{-1}  \>. \label{eq:GenG}
\end{equation}
Remarkably, the Green's function of $\bm{YY}^T$ is the $G_{12}$ entry of the generalized Green's function.
This construction is a slight modification of the quaternionization approach to large non-Hermitian matrices developed by \citep{JNPZ},  therefore we adapt these concepts here for calculations in the large width limit of the network. Furthermore, the generalized Green's function \eqref{eq:GenG} is given implicitly by the solution of the Schwinger-Dyson equation
\begin{equation}
\cG(\cZ)=(\cZ-\cR(\cG(\cZ)))^{-1}. \label{eq:SD1}
\end{equation}
Here $\cR$ is the generalized $R$-transform of FRV theory. This construction is a generalization of standard FRV tools to the matrix-valued functions. In particular, \eqref{eq:SD1} is such a generalization of \eqref{eq:RGrelation} to $2\times 2$ matrices.

To study two common weight initializations, Gaussian and scaled orthogonal, on the same footing, we assume that $W$ belongs to the class of biunitarily invariant random matrices, i.e. its pdf is invariant under multiplication by two orthogonal matrices, $P(\bm{UWV}^T)=P(\bm{W})$ for $\bm{U},\bm{V}\in O(N)$. In the large $N$ limit these matrices are known in free probability as $R$-diagonal operators~\citep{SPEICHER}. A product of $R$-diagonal operator with an arbitrary operator remains $R$-diagonal~\citep{NS}, thus the matrix $\bm{X}$ is $R$-diagonal too.

The generalized $\cR$-transform of $R$-diagonal operators takes a remarkably simple form~\citep{NowTarn}
\begin{equation}
\cR(\cG)=A(G_{12}G_{21})\left(\begin{array}{cc}
0 & G_{12} \\
G_{21} & 0 
\end{array}\right). \label{eq:RRdiag}
\end{equation}
Here, $A(x)=\sum_{k=1}^{\infty} c_{2k} x^{k-1}$ is the determining sequence, which generates cumulants $c_{2k}$, it is a Taylor expansion of $A(x)$ at $0$.  
For the later use we mention a simple relation for the determining sequence of a scaled matrix $A_{aX}(z)=a^2A_X(a^2z)$, which generalizes the Hermitian case $G_{aH}(z)=\frac{1}{a}G_H(\frac{z}{a})$ or, equivalently, 
$R_{aH}(z)=aR_H(az)$. 

We derive the equation for the Green's function (with $G(z)=G_{12}$) by substituting the $\cR$-transform \eqref{eq:RRdiag} into \eqref{eq:SD1} and eliminating irrelevant variables. It thus reads: 
\begin{equation}
G=\frac{GA(zG^2)-1}{a^2-z(1-GA(zG^2))^2},
\end{equation}
where for clarity we omitted the argument of the Green's function. In the next step we substitute $z\to R(z)+\frac{1}{z}$ and use \eqref{eq:RGrelation} to obtain
\begin{equation}
z=\frac{zA(z^2R+z)-1}{a^2-(R+\frac{1}{z})(1-zA(z^2R+z))^2}.
\end{equation}
Then, we substitute $z\to zS(z)$ and use \eqref{eq:RSrelation}, which leads us to
\begin{equation}
1=\frac{zSA(z(z+1)S)-1}{a^2 zS-(z+1)(1-zSA(z(z+1)S))^2}. \label{eq:SEquation}
\end{equation}
This equation is exact.
To incorporate the additional scaling of weights variances by $1/L$ in our considerations, as proposed by~\citep{Taki1} and  \citep{BFLLMM}, we rescale $\bm{X}\to \bm{X}/\sqrt{L}$ and since we are interested in deep networks, we keep only the leading term in $1/L$ (see also~\citep{JW}).
This leads to $A\left(z(z+1)S\right)\to\frac{1}{L}A\left(\frac{1}{L}z(z+1)S\right)=\frac{c_2}{L}+O\left(\frac{1}{L^2}\right)$, which simplifies \eqref{eq:SEquation} to a quadratic equation for $S$. Choosing the appropriate branch of the solution, we see that 
\begin{equation}
S_{Y_lY_l^T}(z)=\frac{1}{a^2}\left(1-\frac{c^l_2}{a^2L}(1+2z)+\mathcal{O}\left(\frac{1}{L^2}\right)\right).
\end{equation}
Here 
\begin{equation}\label{cl2}
c^l_2=\<\frac{1}{N}\tr \bm{W}^l\bm{D}^l\bm{D}^l(\bm{W}^l)^T\>=\frac{\sigma^2_w}{N}\sum_{i}^{N} \left(\phi'(h_i^l)\right)^2
\end{equation}
is the squared spectral radius of the matrix $\bm{D}^l \bm{W}^l$. In general, $c_2^l$ can vary across the depth of the network due to non-constant variance of preactivations. Assuming that this variability is bounded, we can consider the logarithm of \eqref{eq:SProduct} and write:
\begin{equation}
\ln S_{JJ^T}(z)=-2L \ln a -\frac{(1+2z)}{a^2}c, \label{eq:Sapprox}
\end{equation}
where we defined the effective cumulant $c=\frac{1}{L}\sum_{l=1}^{L}c_2^l$ and used $\ln(1+x)\approx x$. This allows us to deduce the form of the $S$-transform, assuming that $a$ does not scale with $L$
\begin{equation}
S_{JJ^T}(z)=\frac{1}{a^{2L}}e^{-\frac{c}{a^2}(1+2z)}. \label{eq:Sproduct}
\end{equation}
Substituting $z\to zR(z)$ and using \eqref{eq:RSrelation}, 
we obtain
\begin{equation}
a^{2L}=R(z)\exp\left[-\frac{c}{a^2}(1+2zR(z))\right].
\end{equation}
Then, we substitute $z\to G(z)$ and use~\eqref{eq:RGrelation} to finally get 
\begin{equation}
a^{2L}G(z)=(zG(z)-1)e^{\frac{c}{a^2}(1-2zG(z))}, \label{eq:GreenEquation}
\end{equation}
an equation for the Green's function characterizing the square singular values of the Jacobian, which can be solved numerically.  
We do that for a range of different activation functions and present the results with numerical simulations to corroborate them in  Fig.~\ref{fig:NumMath}.

 
We close this section with a remark that the above analysis is not restricted only to the model  \eqref{model}, but analogous reasoning can be performed for networks in which skip connections bypass more than one fully connected block.
The qualitative results remain unaltered provided that $L$ is replaced by the number of skip connections.
 

\subsection{Signal propagation}
\label{s:signal}

The formulas we have derived until now were given in terms of a single parameter $c$, which is the squared derivative of the activation function averaged within each layer and across the depth of the network.
 Thus, we now need to address the behavior of preactivations. In the proceeding paragraph, we closely follow a similar derivation done in \citep{SGGS}, for fully connected feed forward networks.

For the simplicity of our arguments, we consider here $W_{ij}^l$ and $b^{l}_{i}$ as  independent identically distributed (iid) Gaussian random variables with $0$ mean and variances $\frac{\left(\sigma_W\right)^2}{LN}$ and $(\sigma_b)^2$, respectively. Here, $(\sigma_W)^2$ is of order one, and the additional scaling is meant to reflect those introduced in the previous paragraphs. At the end of this section we provide an argument that the same results hold for scaled orthogonal matrices. 

In this subsection, we will denote the averaging over variables $W_{ij}^l$ and $b^{l}_{i}$, at a given layer $l$, by $\braket{\cdot}_{wbl}$. By $\<u\>_l$ we denote the sample average, of some variable $u$ in the $l$-th layer: $\braket{u}_l\equiv \frac{1}{N} \sum_{i=1}^{N}u_i^l$. Note that the width ($N$)  is independent of the layer number, however the derivation can be easily generalized to the opposite case, when the architecture is more complicated. Unless stated otherwise explicitly, all integrals are calculated over the real line.

We are interested in the distribution of $h_i^l$ in our model, depending on the input vectors and the probability distributions of $W_{ij}^l$ and $b^l_i$. If we assume they are normal (as can be argued using the Central Limit Theorem), we just need the first two moments. It is clear that $\braket{ h_i^l}_{wbl}=0$. Furthermore, we assume ergodicity, i.e. that averaging some quantity over a layer of neurons is equivalent to averaging this quantity for one neuron over an ensemble of neural networks with random initializations. 
We assume this is true for $h_i$, $x_i$, $W_{ij}$ and $b_i$. Thus, we can say that $\braket{ h}_{l}=0$  and moreover, as we work in the limit of wide networks, $\braket{ f\left(h\right)}_l$ (where $f$ is some function of $h_l$) can be replaced with an averaging over a normal distribution of variance $q^l\equiv\left<\left( h^{l}\right)^2\right>_{wbl}$. This is the crux of the dynamical mean field theory approach \citep{PLRSG} for feed-forward neural networks. We have in particular:

\be 
c_2^l=\sigma^2_W \left<\left(\phi ' (h)\right)^2\right>_l=\sigma^2_W \int\mathcal{D} z \phi '^2\left(\sqrt{q^l}z\right),
\ee
where $\cD z=\exp(-z^2/2) dz/\sqrt{2\pi}$.
To calculate the effective cumulant, we need to know how the variance of the distribution of the preactivations changes as the input information propagates across consecutive layers of the network. It is shown in  Appendix~\ref{sec:SignalDerivation}, that $q^{l}$ satisfy the recurrence equation
\be \label{eq:recursion}
q^{l+1}=a^2  q^l +\left(1-a^2\right) \sigma_b^2 +\frac{(\sigma_W)^2 }{L}\int\mathcal{D} z \phi^2 \left(\sqrt{q^{l}}z\right)+ 
2 \frac{(\sigma_W)^2 }{L}\left[\sum_{k=1}^{l-1} a^k \int \cD z \phi\left(\sqrt{q^{l-k}} z\right)\right]\int\mathcal{D} z \phi \left(\sqrt{q^{l}}z\right),
\ee
with the initial condition $q^1=\sigma_b^2+\frac{\sigma_W^2}{L}$.

We remark here that the above reasoning concerning signal propagation holds also when the weights are scaled orthogonal matrices, i.e. $\bm{WW}^T=\frac{\sigma_W^2}{L}\mathbf{1}$. In such a case 
$\<W_{ij}\>=0$ and $\<W_{ij}W_{kl}\>=\frac{\sigma_W^2}{NL}\delta_{ik}\delta_{jl}$~\citep{CollSn} and the entries of $\bm{W}$ can be approximated as independent Gaussians~\citep{Chatterjee}.

\subsection{Random matrix simulations}\label{s:rmt_exp}

\begin{figure}[t]\begin{center}
\includegraphics[width=0.6\textwidth]{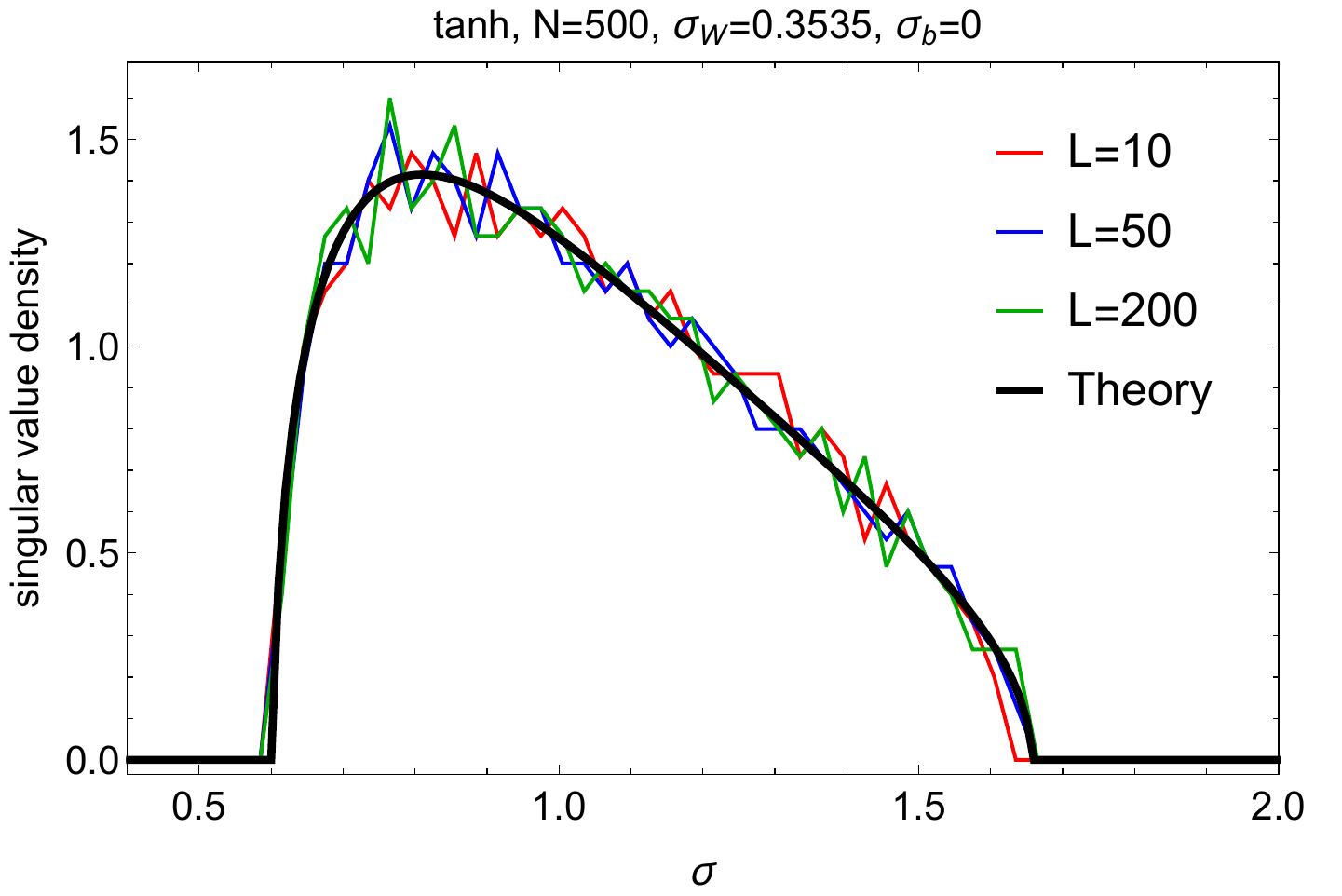} \\
\includegraphics[width=0.6\textwidth]{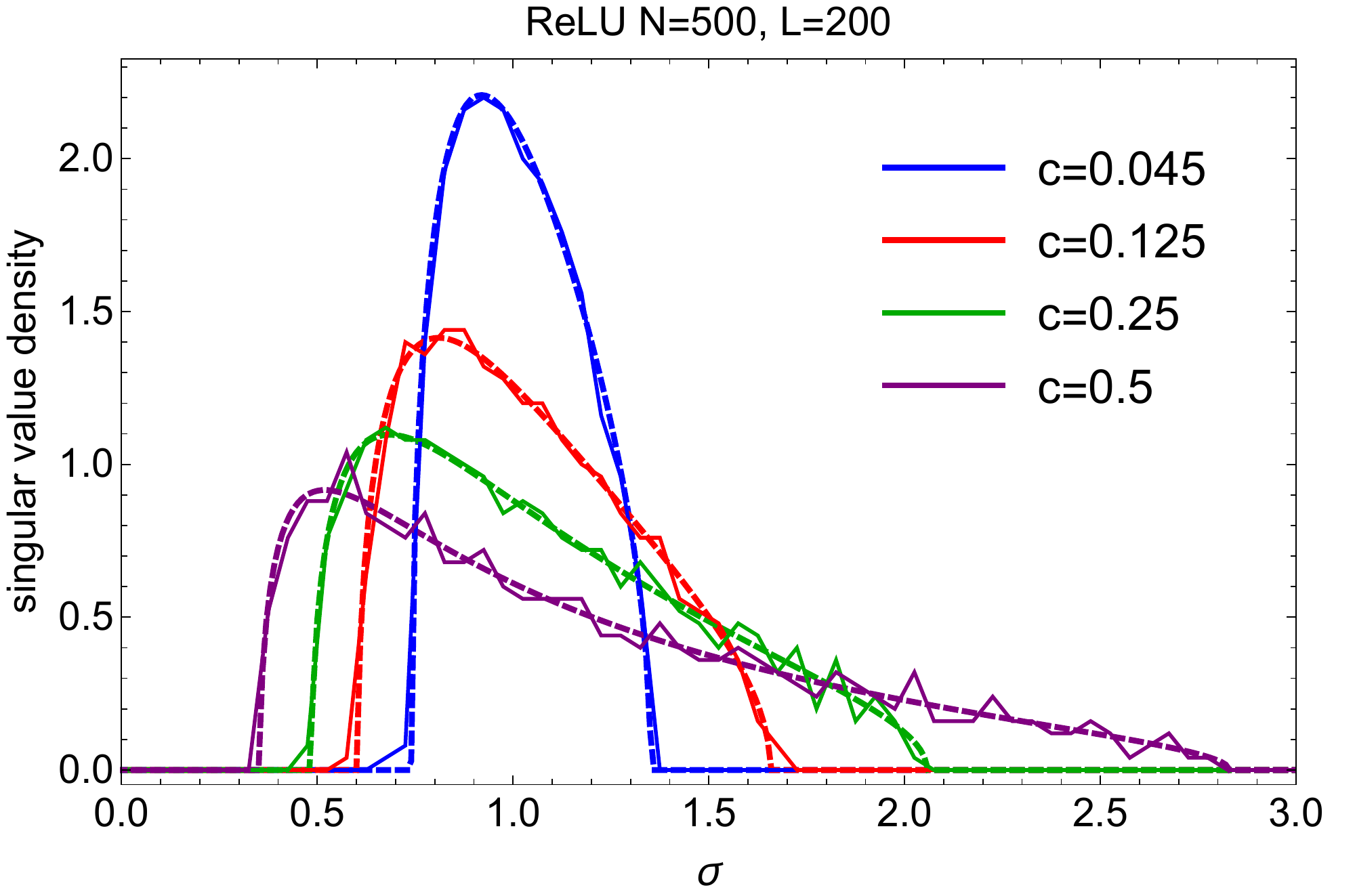} 
\end{center}
\caption{(Top) density of singular values of the input-output Jacobian for the residual network with tanh nonlinearity. Note that the asymptotic theoretical result describes remarkably well not very deep ($L=10$) networks. (Bottom) asymptotic distribution of singular values for various values of parameter $c$ (dashed) juxtaposed with the numerical simulations for ReLU nonlinearity (solid). Note that  histograms were calculated from a single random initialisation. The smaller $c$, the narrower the spectrum and the closer to the ideal isometry.
\label{fig:NumMath}
}
\end{figure}

To thoroughly test the theoretical predictions of Section \ref{spectral}, we run numerical simulations using \textit{Mathematica}. 
The initial condition, input vector $\bm{x}^0$ of length $N=500$, filled with iid Gaussian random variables of zero mean and unit variance, is propagated according to the recurrence \eqref{model}, for various activation functions. The network weights and biases are generated from normal distribution of zero mean and $\sigma_W^2/NL$ and $\sigma_b^2$ variances, respectively, with $N=500$ 
The propagation of variance of pre-activations, post-activations as well as the calculation of the second cumulants $c^l$ for the studied activation functions, across the network, is presented in Appendix~\ref{a:1}. All numerical simulations corroborate our theoretical results. Here, for clarity and as a generic example, in Fig. \ref{fig:NumMath} (upper), we show the distribution of singular values of the input-output Jacobian (defined in~\eqref{eq:jac1}) for the tanh nonlinearity for various network depths. In this example the Jacobian in not independent of the signal propagation, contrary to the case of piecewise linear activation functions.  
Similarly, in the lower panel of Fig. \ref{fig:NumMath}, we showcase the outcome of numerical experiments and the associated, matching, theoretical results for the most popular ReLU activation function, for various initializations resulting in different values of the effective cumulant $c$.

\section{Experiments on image classification}
\label{s:Experiments}

\begin{figure}
\begin{center}
\includegraphics[scale=0.25]{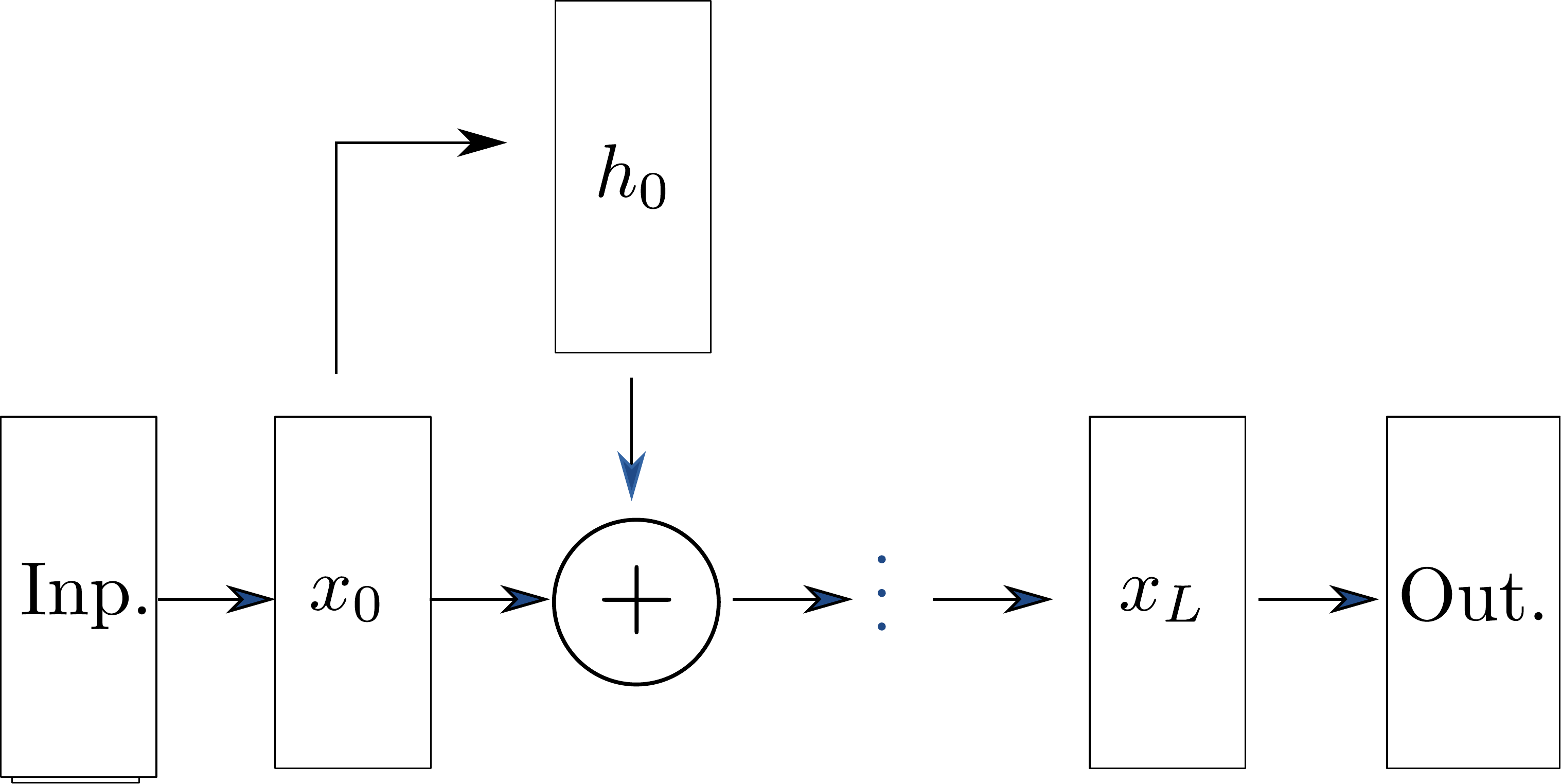}
\end{center}
\caption{Residual network architecture used in the paper.\label{fig:resnet}}
\end{figure}  

The goal of this section is to test our theoretical predictions on real data via the popular CIFAR-10 benchmark~\citep{CIFAR10}. To this end we will use a single representation fully connected residual network, see Fig.~\ref{fig:resnet}. This simplified version of the model of~\citep{HZRS} does not use (i) multiple stages with different dimension of hidden representation, and (ii) two layers within residual block. We leave study of a more general version of ResNets for future work.

\subsection{Achieving dynamical isometry for any activation function} \label{s:numerics1}

Perhaps the most interesting prediction of our theory is that ResNets, in contrast to fully connected networks, can achieve dynamical isometry for many different activation functions. We will study this empirically by looking at $\mathbf{J}$, at initialization, for different activation functions and number of residual blocks. Please note that by $\mathbf{J}$ we refer to Jacobian of the output of the last residual block with respect to the input of the first one, see also Fig.~\ref{fig:resnet}.

We consider the following popular activation functions: ReLU~\citep{NH}, Tanh, Hard Tanh, Sigmoid, SeLU~\citep{KUM} and Leaky ReLU~\citep{MHN} with the leaking constant $0.05$ and $0.25$. For each activation function we consider the number of blocks $L$ to be $10$ and $20$. All weights of the network are initialized from a zero-centered normal distribution whereas biases are initialized to zero. The weights of the residual blocks are initialized using standard deviation $\sigma_W/\sqrt{NL}$, other weights are initialized as by~\citep{Taki1}. For the given activation function and the number of blocks $L$, we set $\sigma_W$ in such a way that the effective cumulant $c=0.125$, which ensures the concentration of eigenvalues of the Jacobian around one, and hence dynamical isometry (see  Appendix~\ref{s:actfunc} for more details and Fig. \ref{fig:NumMath} for the shape of the singular value densities).

\begin{figure}
\begin{center}
\includegraphics[width=0.7\textwidth]{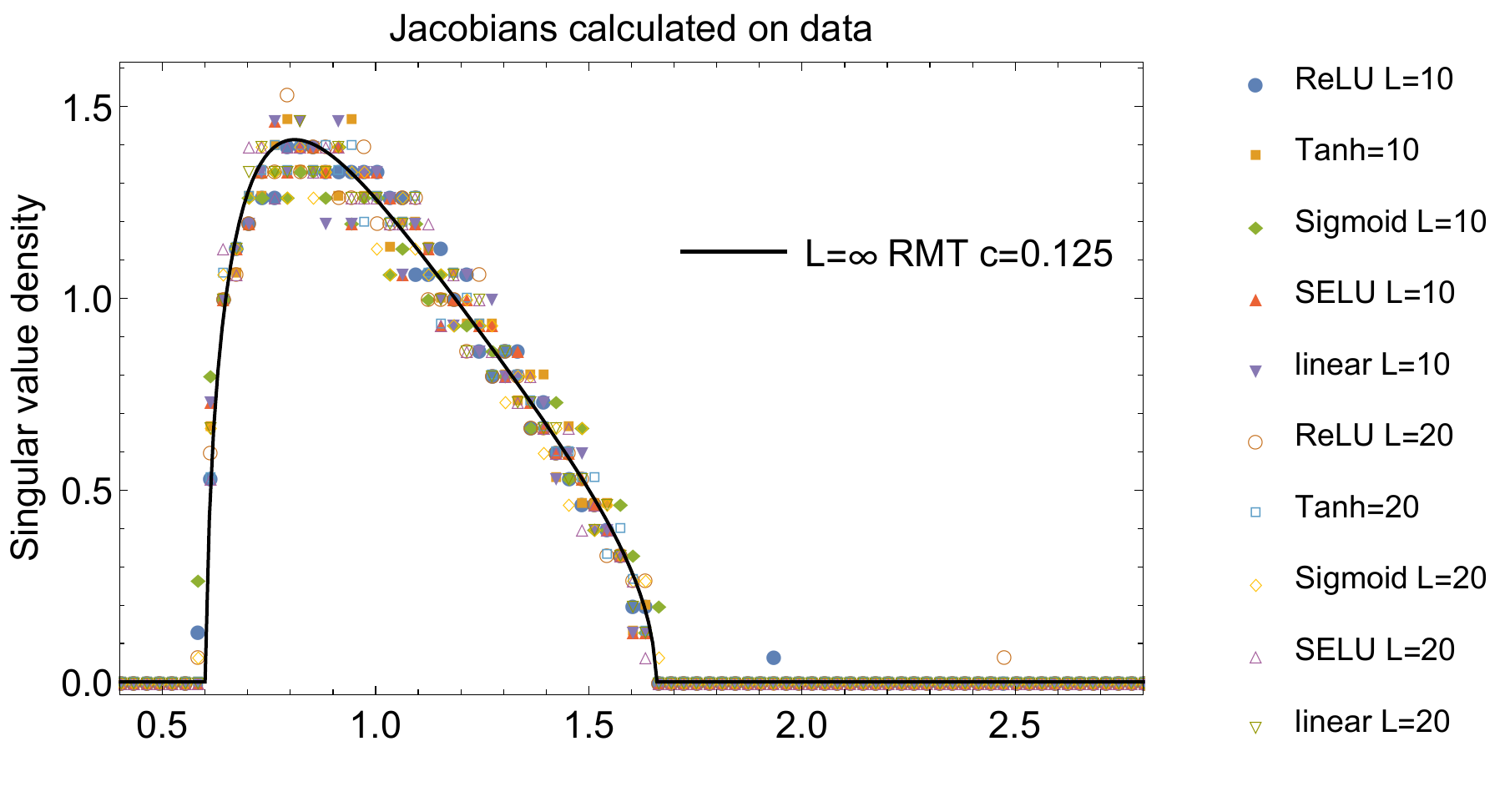}
\end{center}
\caption{Singular spectra obtained for various activation functions and depth $L=10,20$. The network was fed with examples from CIFAR10 dataset. \label{fig:JacSpectra}}
\end{figure}

For each pair of activation function and number of blocks we compute the empirical spectrum of $\mathbf{J}$ at initialization, the results are reported in Fig.~\ref{fig:JacSpectra}. Indeed, we observe that upon scaling the initializations standard deviation, in such a way that $c$ is kept constant, the empirical spectrum of $\mathbf{J}$ is independent of the number of residual blocks or the choice of activation functions.

\subsection{Learning dynamics are more similar at universality under dynamical isometry}\label{s:nnexp}

\begin{figure}\begin{center}
\includegraphics[width=0.48\textwidth]{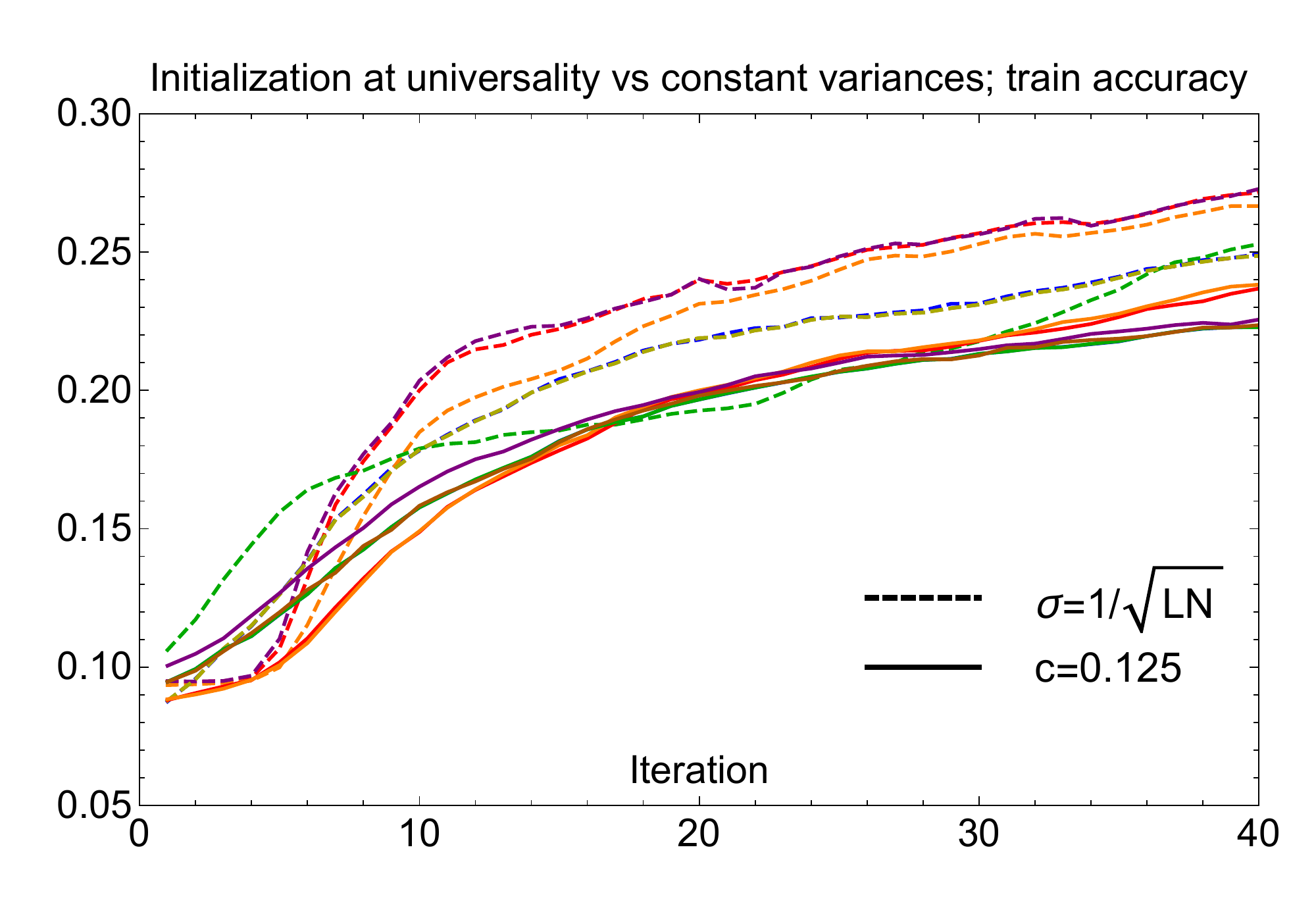}
\includegraphics[width=0.48\textwidth]{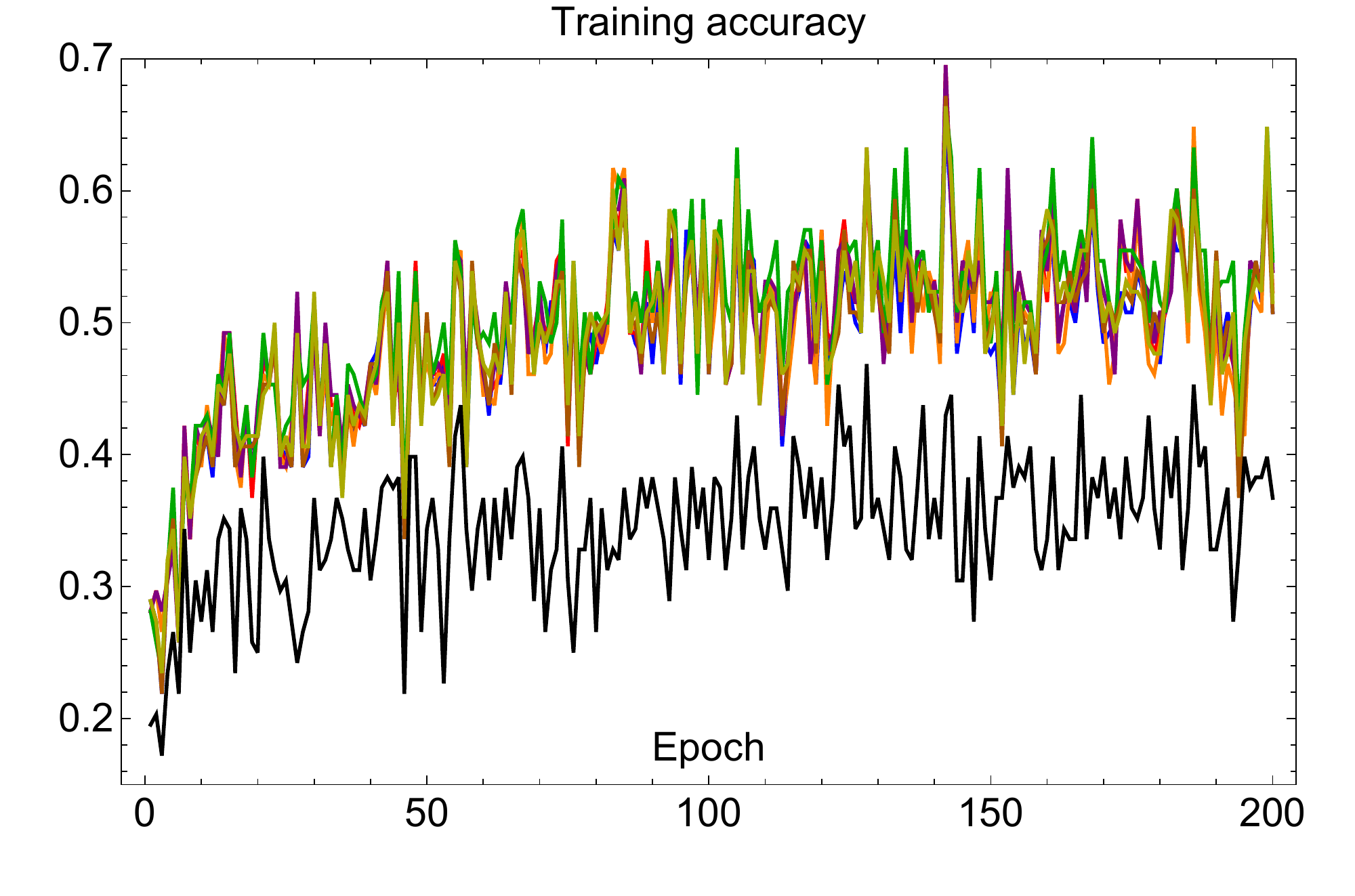}\\
\includegraphics[width=0.65\textwidth]{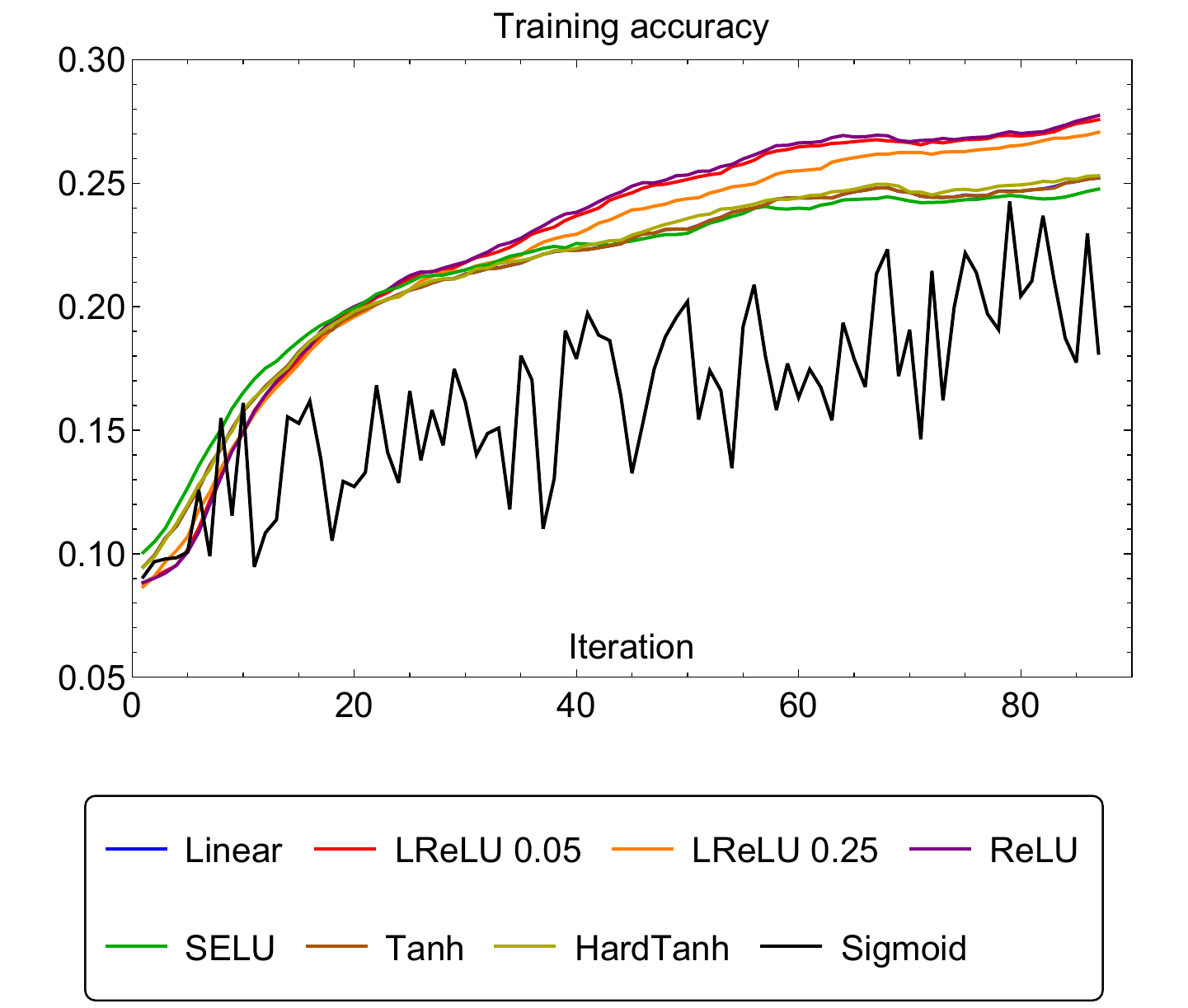} 
\end{center}
\caption{Training accuracy during first 200 epochs (middle) and first 100 iterations (bottom) of residual networks with various activation functions. The weight initialization was chosen for each activation function in such a way that the effective cumulant is $c=0.125$. In the top panel, the dynamics with this initialization was juxtaposed with analogous training of networks in which the variance of weights was chosen to be $\frac{1}{LN}$ for all activation functions.  We used leaky ReLU with $\alpha=0.05, 0.25$. \label{fig:TrainAcc}  }
\end{figure}

Our next objective is to investigate whether networks achieving dynamical isometry share similar learning dynamics. While this is outside of the scope of our theoretical investigation, it is inspired by studies such as \citep{PSG1}, which demonstrate the importance of dynamical isometry at the initialization for the subsequent optimization.

We consider the same set of experiments as in the previous section, and follow a similar training protocol to~\citep{HZRS}. We train for $200$ epochs and drop the learning rate by a factor of $10$ after epochs $80$, $120$, and $160$. We use batch-size $128$ and a starting learning rate of either $10^{-3}$ or $10^{-4}$
\footnote{We use relatively low learning rates, largely because we omit batch normalization layers in the architecture.}.

First, we look at the learning dynamics on the training set. We can observe that most of the activation functions exhibit similar training accuracy evolution, see Fig.~\ref{fig:TrainAcc} (middle). Using the sigmoid activation, led however to significantly slower optimization. 
This is due to a faster growth of the variances of post- and pre- activations (which can be observed in Fig. \ref{Fig:Verification2}), which exacerbates the neuron saturation problem.

Overall our results suggest that the singular spectrum of $\mathbf{J}$ at initialization does not fully determine generalization and training performance. Nonetheless, setting the same effective cumulant
for the experiments with different activation functions 
results in a markedly coinciding behavior of neural networks using activation functions of similar characteristics.
This is in contrast to a setup in which the variances of the weight matrix entries are set to be equal. To demonstrate this we run another set of training experiments, this time with all standard deviations $\sigma=1/\sqrt{LN}$. The plots depicting the full results are relegated to Fig. \ref{Fig:baseline} in Appendix \ref{a:2}.
Here, in Fig.~\ref{fig:TrainAcc} (top) we showcase the training accuracy during the first $40$ iterations for these two setups (excluding, for clarity, the networks with the sigmoid activation function). With different effective cumulants, the network learning dynamics, differs among experiments with different activation functions, especially at the beginning of learning. 

This suggests that the spectrum of the input-output Jacobian at initialization can be treated as a confounding variable in such experiments. Ensuring that the level of dynamic isometry, and hence the value of the effective cumulant is kept the same, provides the possibility of a more meaningful comparison of the effect of activation functions on learning dynamics.

\section{Comment on batch normalization}\label{s:bn}

A crucial component of practically used ResNets is batch normalization \citep{IS}. When it is used on pre-activations, between each layer, the propagation of the information in the network is described by:
\be  \label{model_withbn}
\bm{x}^l=\phi(\bm{y}^l)+a \bm{x}^{l-1}, \,\,\,\, \bm{h}^l=\bm{W}^l \bm{x}^{l-1}+\bm{b}^l, \ee
with 
\be
y_{i}^{l}= \frac{h_{i}^{l}-\mu_{i}^{l}}{\sigma_{i}^{l}} \gamma_{i}^{l}+ \beta_{i}^{l},
\ee
where 
$\mu_i^l$ is the mean and $\sigma_i^l$ (regularized with some small $\epsilon$) is the standard deviation of the $k$'th mini batch inputs $i$'th coefficient in layer $l$. $\gamma_{i}^{l}$ and $\beta_{i}^{l}$ are parameters optimized during the learning process. 
In this case, the formula for the Jacobian reads:
\be 
\bm{J}=\prod_{l=1}^{L}\left(\bm{D}^l {\bm H}^{l} \bm{W}^l +{\bm 1}a\right),
\ee
where $\bm{H}^l$ is a diagonal matrix such that $H^l_{ij}= \delta_{ij}\, \gamma_i^l / \sigma^l_i$.
Therefore, the only difference in the spectral statistics derivation from the previous section, is that (\ref{cl2}) becomes
\begin{equation}\label{cl2bn}
c^l_{2{\rm BN}}=\sigma^2_W \left<\left(\frac{\gamma^l }{\sigma^l} \phi'(y^l)\right)^2\right>_l.
\end{equation}
Thus, the universal, large $L$ limit equation for the Green's function of the Jacobian (\ref{eq:GreenEquation}) 
holds also when batch normalization is included. Again, $\sigma^l_i$ and $y^l_i$ can be treated as random variables. Unfortunately, the evolution of their probability density functions across the layers is more complicated and beyond the scope of this paper.

\section{Synopsis and discussion}
The main focus of this paper was the singular spectrum of the input-output Jacobian of a simple model of residual neural networks. We have shown that in the large network depth limit, it is described by a single, universal equation. This holds irrespective of the activation function used, for biunitarily invariant weight initializations matrices, a set covering Gaussian and scaled orthogonal initialization schemes. The  singular value density  depends on a single parameter called the effective cumulant, which can be calculated by considering the propagation of information in the network, via a dynamical mean field theory approach. This  parameter depends on the activation function used, variance of biases and the entries of the weight matrices, and, for some activation functions, also on the depth of the network.  We demonstrated the validity of our theoretical results in numerical experiments, both by generating random matrices adhering to the assumptions of the model and by evaluating the Jacobians of residual networks (at initialization) on the CIFAR10 dataset. 

For a given activation function and/or network depth, it is always possible to set the weight matrix entries variances in such a way, that the resulting singular spectra of the Jacobians not only fulfill the conditions for dynamical isometry, but also are exactly the same, irrespective of the activation function used. This observation allows us to eliminate the singular spectrum of the Jacobian treated as a confounding factor in experiments with the learning process of simple residual neural networks for different activation functions. 
As an example of how this approach can be applied, we examined how accuracies of simple residual neural networks, employing a variety of activation functions, change during the learning process. When using the same variances of weight matrices entries, the learning curves of similar activation functions differed between each other more than when the networks were initialized with the same input-output Jacobian spectra. This allows, in our opinion, for a more meaningful comparison between the effects of choosing the activation function. We hope this observation will help with the research of deep neural networks. 

 \subsubsection*{Acknowledgements}

PW would like to acknowledge the funding of the Polish National Science Centre, through the project SONATA number 2016/21/D/ST2/01142. WT appreciates the financial support from the Polish Ministry of Science and Higher Education through ''Diamond Grant'' 0225/DIA/2015/44. SJ was supported by the Polish National Science Centre through ETIUDA stipend No.~2017/24/T/ST6/00487.

\bibliography{refs}

\newpage
\appendix
\section*{Appendices}

\section{Spectrum of the Jacobian}
\label{sec:SpectralEdges}
To make the characteristics of the spectrum more explicit, we shall calculate the positions of the spectral edges of the probability density of squared singular values. Their locations, $z*$, can be determined from the condition $\frac{1}{G'(z*)}=0$. In this case we assume $a=1$ for simplicity and take a derivative of \eqref{eq:GreenEquation}, obtaining
\begin{equation}
G'=(zG'+G)e^{c(1-2zG)}-(zG-1)e^{c(1-2zG)}2c(G+zG').
\end{equation}
The exponent can be eliminated with the help of \eqref{eq:GreenEquation}, leading to
\begin{equation}
1=\left(z+\frac{G}{G'}\right)\frac{G}{zG-1}-2cG\left(\frac{G}{G'}+z\right).
\end{equation}
Taking $\frac{1}{G'}=0$, we arrive to the quadratic equation
\begin{equation}
2cz^2G^2-2c zG-1=0,
\end{equation}
which, together with \eqref{eq:GreenEquation}, determine the location of the edges and the value of the Green's function at these points. Solving, we obtain
\begin{equation}
z_{\pm}=\left(1+c\pm\sqrt{c(2+c)}\right)e^{\pm\sqrt{c(2+c)}}.
\end{equation}
Note that the perfect isometry (i.e. all eigenvalues are 1) is achieved for $c=0$, as independently proposed by \citep{OR}, while for small $c$ the size of the support grows sublinearly $z_{\pm}\approx 1\pm2\sqrt{2c}$. Moreover, for large $c$, that is far from dynamical isometry, the largest eigenvalue is exponentially large, while the smallest is exponentially small. This fact underscores importance of a proper initialization.


\section{Detailed derivation of the signal propagation}
\label{sec:SignalDerivation}
With the scalings of  from section \ref{s:spectrala} made explicit, we have $\left< W_{ij}^lW_{km}^{l}\right>_{wbl}=\left< W_{ij}^l W_{im}^{l}\right>_{l}=\delta_{ik}\delta_{jm}\frac{(\sigma_W)^2}{ LN}$.
Now, based on (\ref{model}) and the above considerations, we have
\be \label{eq:h2}
q^l=\frac{(\sigma_W)^2 }{L} \left< x^2\right>_{l-1}+(\sigma_b)^2
\ee
and
\be \label{eq:x2}
\left< x^2\right>_{l-1}=\left<\phi\left( h^{l-1}\right)^2\right>_{l-1}+\frac{2a}{N}\sum_{i=1}^{N}\phi\left( h_i^{l-1}\right)x_i^{l-2}+a^2 \left< x^2\right>_{l-2}
\ee
We assume that the factorization $\frac{1}{N}\sum_{k} x_k^{l-1}\phi(h^l_k)=\<x\>_{l-1} \int \cD u \phi(u\sqrt{q^l})$ holds in the large $N$ limit. This is justified, as the input to $h^l_k$ comes from all the many elements of $x^{l-1}$. We can rewrite (\ref{eq:x2}) as 
\be \label{eq:x2_1}
\left< x^2\right>_{l-1}=\int\mathcal{D} z \phi^2 \left(\sqrt{q^{l-1}}z\right)+ 
2a \left< x\right>_{l-2} \int\mathcal{D} z \phi \left(\sqrt{q^{l-1}}z\right) +a^2  \left< x^2\right>_{l-2}
\ee
Turning to $\left< x\right>_{l-2}$, based on (\ref{model}), we have:
\begin{equation}
\<x\>_l=a \<x\>_{l-1}+\int \cD z\phi\left(\sqrt{q^l}z\right).
\end{equation}
For $\<x\>_0=0$ the recurrence yields
\begin{equation}
\<x\>_l= \sum_{k=0}^{l-1} a^k \int \cD z \phi\left(\sqrt{q^{l-k}} z\right).
\end{equation}
Thus, (\ref{eq:x2_1}), with a shift in $l$, turns into
\be \label{eq:x2_2}
\left< x^2\right>_{l}=\int\mathcal{D} z \phi^2 \left(\sqrt{q^{l}}z\right)+ 
2\left[\sum_{k=1}^{l-1} a^k \int \cD z \phi\left(\sqrt{q^{l-k}} z\right)\right]\int\mathcal{D} z \phi \left(\sqrt{q^{l}}z\right) +a^2  \left< x^2\right>_{l-1}.
\ee

Finally, we use (\ref{eq:h2}) to obtain 
\be \label{eq:recursion}
q^{l+1}=a^2  q^l +\left(1-a^2\right) \sigma_b^2 +\frac{(\sigma_W)^2 }{L}\int\mathcal{D} z \phi^2 \left(\sqrt{q^{l}}z\right)+ 
2 \frac{(\sigma_W)^2 }{L}\left[\sum_{k=1}^{l-1} a^k \int \cD z \phi\left(\sqrt{q^{l-k}} z\right)\right]\int\mathcal{D} z \phi \left(\sqrt{q^{l}}z\right),
\ee
which is a closed recursive equation for $q^l$. We note that for $a=0$, the known, feed-forward network recursion relation is recovered. Furthermore, for the case of $a=1$, in contrast to the feed-forward architecture, the biases do not influence the statistical properties of the pre-activations. Moreover, for ResNets, this recursive relation is iteratively additive, namely each $q^{l+1}$ is a result of adding some terms to the previous $q^l$. In all the examples studied below, the first term is positive and the second term is non-negative.
This in turn means that the variance of pre-activations grows with the networks depth and there are no non-trivial fixed points of this recursion equation. Finally, here we can see the importance of the $\frac 1N$ scaling of $(\sigma_W)^2$, without which, $q^l$ would grow uncontrollably with $l$.


\section{Results for various activation functions}\label{s:actfunc}
We now investigate particular examples of activation functions. For simplicity, we consider purely residual networks (we set $a=1$). The numerical verifications of the results presented here will follow in the next subsection.
\begin{enumerate}
\item Linear

In the case of the linear activation function $\phi'(x)=1$ and there is no need to consider the way the pre-activations change across the network. Thus we can proceed to calculating the cumulant which yields $c=c_2=\sigma_W^2$. 
\item Rectified Linear Unit

The example of ReLU is only slightly more involved. Now we have $\phi'(x)=\theta(x)$, where $\theta(x)$ is the Heaviside theta function, and thus
\begin{equation}
c_2^l=\sigma_W^2\int\cD u \phi'^2\left(u \sqrt{q^l}\right)=\int_{0}^{\infty}\cD u =\frac{1}{2} \sigma_W^2.
\end{equation}

 \item Leaky ReLU
 
 The activation function interpolating between the first two examples is  $\phi(x)=\max(\alpha x,x)$ with $0<\alpha <1$. In this case
\begin{equation}
c_2^l=\sigma_W^2\left( \int_{-\infty}^{0}\alpha^2 \cD u+\int_{0}^{\infty}\cD u \right)=\frac{\sigma_W^2}{2}(1+\alpha^2).
\end{equation}
\end{enumerate}
All together, this leads to the following equation for the Green's function
\begin{equation}\label{Greensfinal_1}
G(z)=\left(zG(z)-1\right)e^{\frac 12 \sigma_W^2(1+\alpha^2)(1-2zG(z))},
\end{equation}
where $\alpha=1$ corresponds to the linear activation function and $\alpha=0$ to ReLU.

Equation (\ref{Greensfinal_1}) can be easily solved numerically for the spectral probability density of the Jacobian. For completeness, we write down the recursion relation (\ref{eq:recursion}) in these three cases:
\be 
q^{l}=q^{l-1} +\frac{\sigma_W^2}{2L}\left(\alpha^2+1\right)q^{l-1}+\frac{\sigma_W^2}{\pi L}(1-\alpha)^2 \sqrt{q^{l-1}}\left(\sum_{k=1}^{l-2}\sqrt{q^k}\right).
\ee 
For the linear activation function ($\alpha=1$), its solution is readily available and reads
\be 
q^l=q^1\left(1+\frac{\sigma_W^2}{L}\right)^{l-1}\simeq q^1 e^{(l-1)\frac{ \sigma_W^2}{L}},
\ee
which explicitly shows the importance of the $1/L$ rescaling introduced earlier.
\begin{enumerate}
\setcounter{enumi}{3}

 \item Hard  hyperbolic tangent
 
The hard tanh activation function is defined by $\phi(x)=x$ for $|x|\leq 1$ and $\phi(x)={\rm sgn(x)}$ elsewhere. Thus:
\be 
c_2^l= \sigma^2_W \int^1_{-1}\cD u = \sigma^2_W \,{\rm erf}\left(\frac{1}{\sqrt{2}}\right)=c .
\ee 
The resulting recurrence  equation for the variance of the preactivations reads:
\be 
q^{l+1}=q^l\left[1+\frac{\sigma^2_W}{L}\left(\,{\rm erf}\left(\frac{1}{\sqrt{2}}\right)-\sqrt{\frac{2}{\pi e}}\right)\right]+ \frac{\sigma^2_W}{L}\left(1-\,{\rm erf}\left(\frac{1}{\sqrt{2}}\right)\right)
\ee
and can be easily solved.

\end{enumerate}
In the preceding examples we dealt with piecewise linear activation functions. Note that in these cases the parameter $c$ does not depend on the variance of biases and linearly increases with the variance of weights. For other nonlinear activation functions to obtain the cumulants, we need to use the recurrence relation describing the signal propagation in the network.

\begin{enumerate}
\setcounter{enumi}{4}
\item Hyperbolic tangent  

When $\phi(x)=\tanh(x)$, the activation function is antisymmetric and the last term of (\ref{eq:recursion}) vanishes. Thus the recurrence takes the form:
\begin{eqnarray}
q^{l+1}= q^l +\frac{(\sigma_W)^2 }{L}\int\mathcal{D} z \phi^2 \left(\sqrt{q^{l}}z\right)
\end{eqnarray}
In the large $L$ limit, we can write
\begin{eqnarray}
q^{l+1}= q^l +\frac{(\sigma_W)^2 }{L}\int\mathcal{D} z \phi^2 \left(\sqrt{q^{l-1}+\Delta^l}z\right),
\end{eqnarray}
where we assume $\Delta^l \sim 1/L$. Expanding this recursively around $q^{l-1}$ for decreasing $l$ and keeping only the leading term, as long as for any $k$, $q^k \gg 1/L^2$, we obtain :
\begin{eqnarray}
q^{l+1}\approx q^l +\frac{(\sigma_W)^2 }{L}\int\mathcal{D} z \phi^2 \left(\sqrt{q^{1}}z\right).
\end{eqnarray}
Therefore, the solution of the recursion is:
\begin{eqnarray}
 q^l \approx q^1+ (l-1) \frac{(\sigma_W)^2 }{L}\int\mathcal{D} z \phi^2 \left(\sqrt{q^{1}}z\right). \label{eq:Approx}
\end{eqnarray}
The variance grows linearly with $l$ (this is verified in appendix \ref{a:1}, see Fig.~\ref{Fig:Verification2}).
Cumulants $c^l_2$ and thus $c$ are obtained with numerical integration.

\end{enumerate}
In fact, in the above calculations we have only used the antisymmetry  property of the hyperbolic tangent activation function and the properties of its behavior near $q^1$. Therefore, these results are valid for other antisymmetric activation functions like ${\rm \phi(x)=arctan(x)}$.

\begin{enumerate}
\setcounter{enumi}{5}

\item Sigmoid 

The sigmoid activation function, $\phi(x)=\frac{1}{1+e^{-x}}$ is the first example we encounter, for which $\<\phi(h)\>\neq 0$, thus it deserves special attention. In particular, in this case one needs to additionally address the last term in (\ref{eq:recursion}). It turns out, that 
\begin{equation}
\int \cD z \phi\left(\sqrt{q^l} z\right)=\frac{1}{2} \label{eq:MeanSigmoid}
\end{equation}
irrespective of $l$. Therefore, the recurrence relation becomes:
\be \label{eq:recursion_sigmoid}
q^{l+1}= q^l+\frac{(\sigma_W)^2 }{L}\int\mathcal{D} z \phi^2 \left(\sqrt{q^{l}}z\right)+  \frac{(\sigma_W)^2 }{2L}(l-1).
\ee
Thus, we can see, that due to the non-zero first moment of the activation function \eqref{eq:MeanSigmoid} the mean and the second moment of post-activations grow with depth. Similarly, the variance of pre-activations increases as the signal propagates, which causes quick saturation of the sigmoid nonlinearity. This in turn precludes training of deep networks~\citep{GB}.

Analogically to the previous case, one can derive an approximation to the solution of the recursion relation. In this case it becomes:
\be \label{eq:recursion_sigmoid_approx}
q^{l+1}\approx q^1+\frac{(\sigma_W)^2 l}{L}\int\mathcal{D} z \phi^2 \left(\sqrt{q^{1}}z\right)+  \frac{(\sigma_W)^2 }{4L}l(l-1).
\ee
We verify this result in Fig.~\ref{Fig:Verification2} in Appendix~\ref{a:1}

\item Scaled Exponential Linear Units

Our final example is the SELU activation function, one introduced recently in \citep{KUM} with the intent to bypass the batch normalization procedure. In this case, we have  $\phi(x)=\lambda x $ for $x>0$ and $\phi(x)=\lambda \beta \left(e^x -1\right) $ for $x\le 0$. Thus, it is not antisymmetric and is nonlinear for negative arguments.
It turns out, that
\be 
c^l_2= \frac{(\sigma_W)^2 \lambda^2}{2}\left[1+\beta^2e^{2q^l} \,{\rm erfc}\left(\sqrt{2 q^l}\right)\right].
\ee
Moreover:
\be 
\int\mathcal{D} z \phi^2 \left(\sqrt{q^{l}}z\right) = \frac{\lambda^2 q^l}2 + 
\frac{\beta^2 \lambda^2}{2}\left[1+e^{2q^l} \,{\rm erfc}\left(\sqrt{2 q^l}\right)-2e^{q^l/2} \,{\rm erfc}\left(\sqrt{\frac{q^l}2}\right)\right]
\ee
and
\be 
\int\mathcal{D} z \phi \left(\sqrt{q^{l}}z\right) = \lambda\sqrt{\frac{q^l}{2\pi}} +\frac{\lambda\beta}{2}\left(e^{q^l/2}\rm{erfc}\left(\sqrt{\frac{q^l}{2}}\right)-1\right). 
\ee
These yield the recursion relation for $q^l$ via (\ref{eq:recursion}). One can check that for $\beta=0$ and $\lambda=1$, the results for ReLU are recovered.
\end{enumerate}

These theoretical predictions for the recursion relations are tested with numerical simulations using Mathematica. The results are relegated to Appendix \ref{a:1}.

\section{Numerical verification of the recurrence relations} \label{a:1}
\begin{figure}[h]
\includegraphics[width=0.33\textwidth]{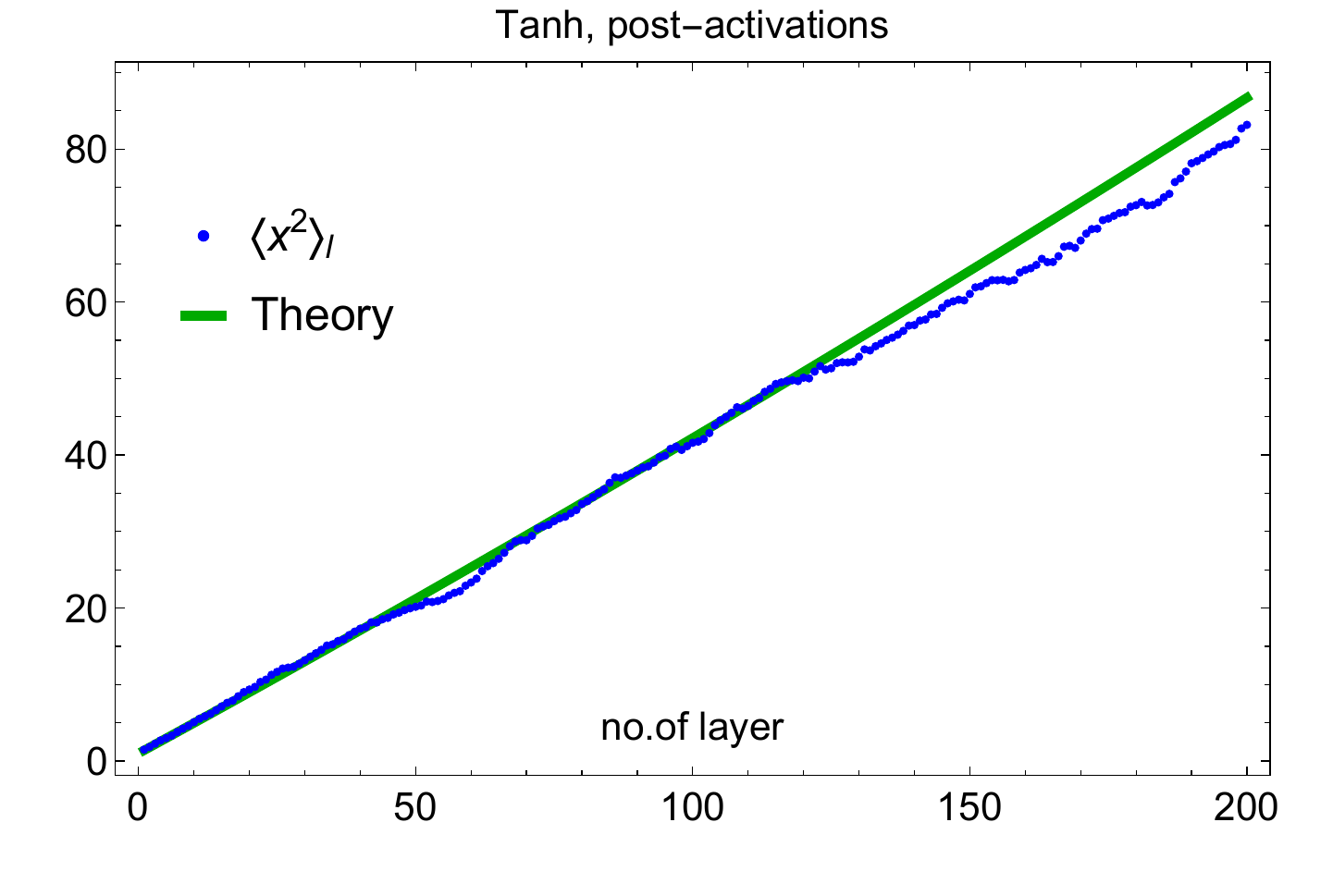}\includegraphics[width=0.33\textwidth]{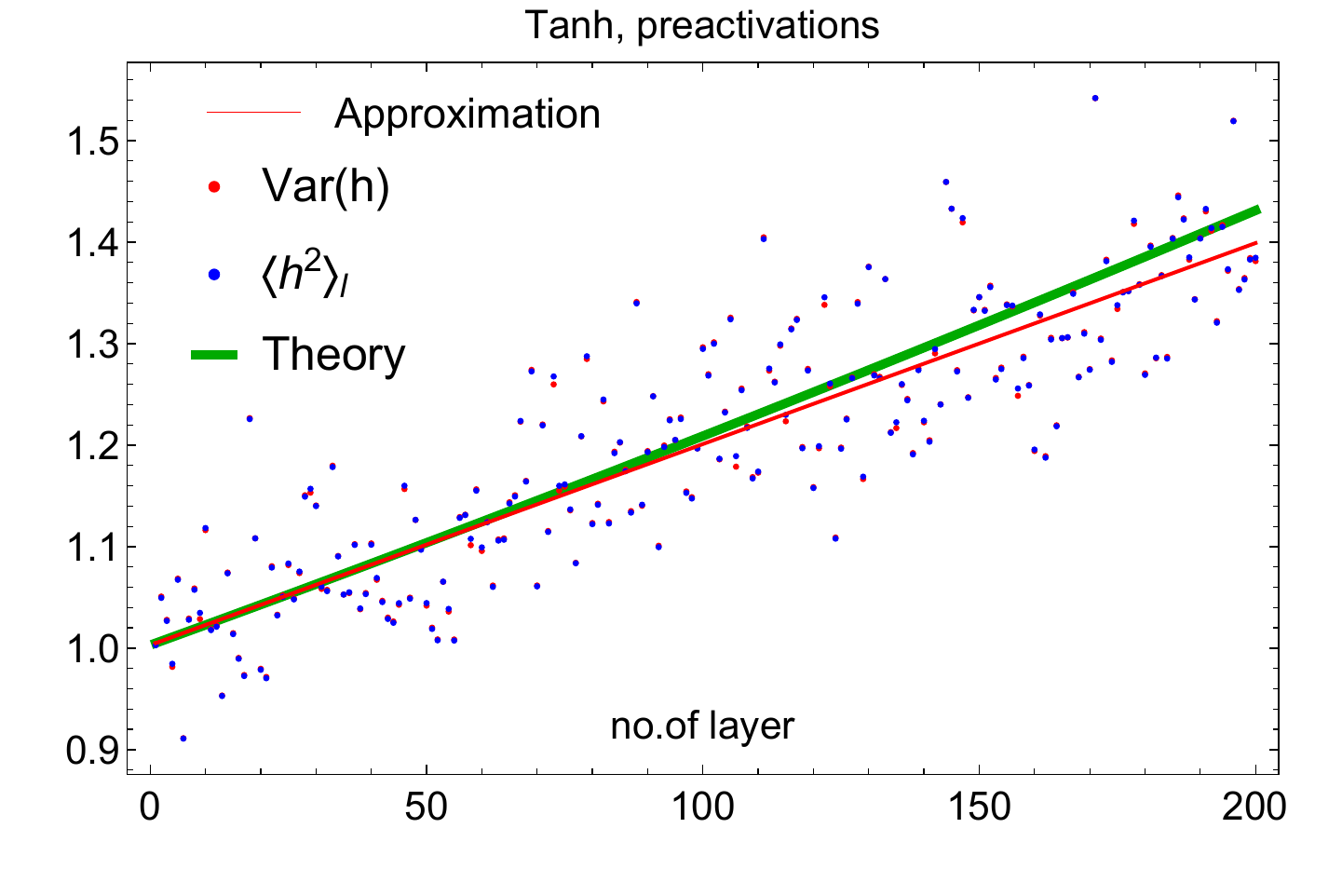}\includegraphics[width=0.33\textwidth]{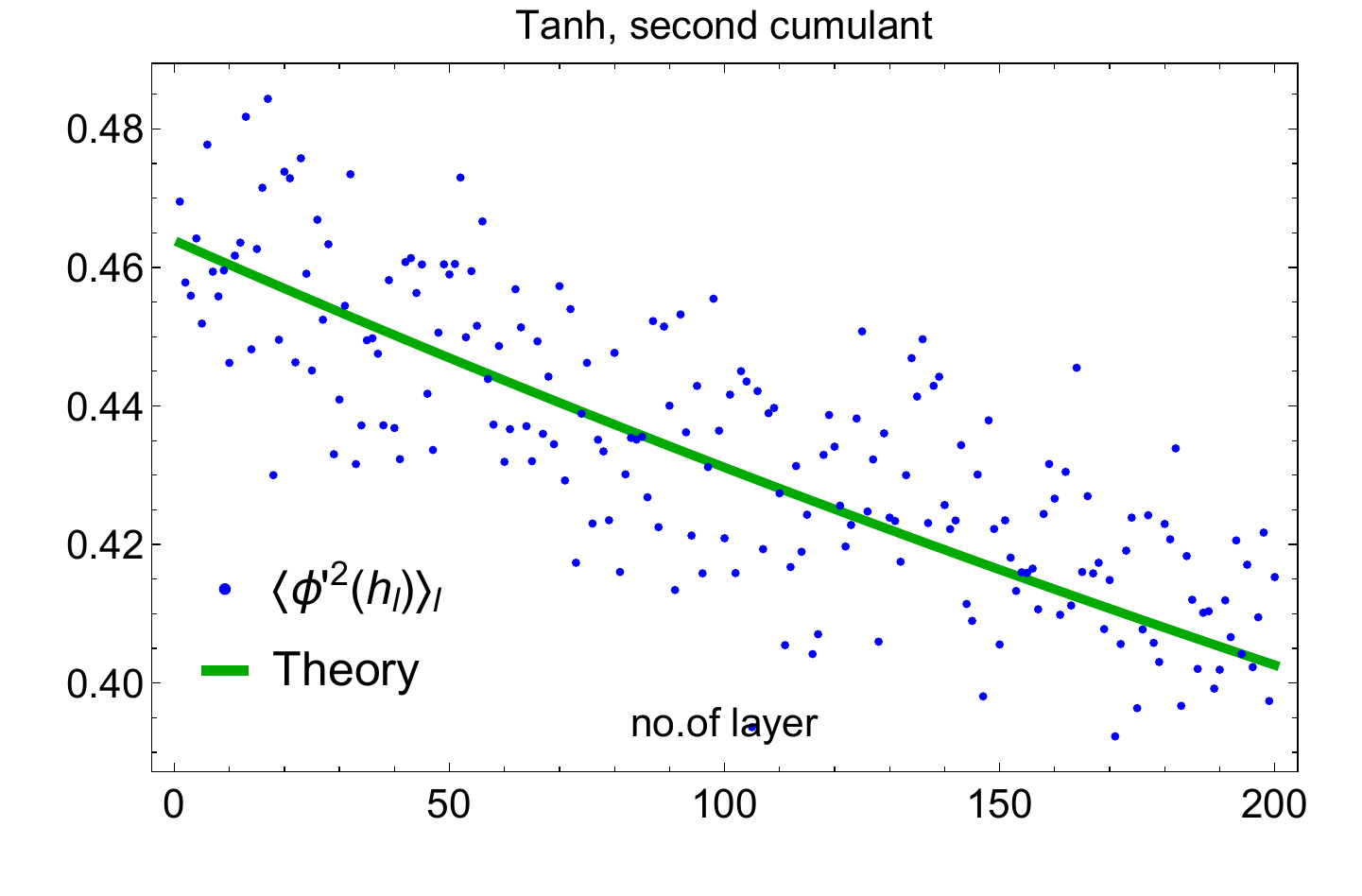} \\
\includegraphics[width=0.33\textwidth]{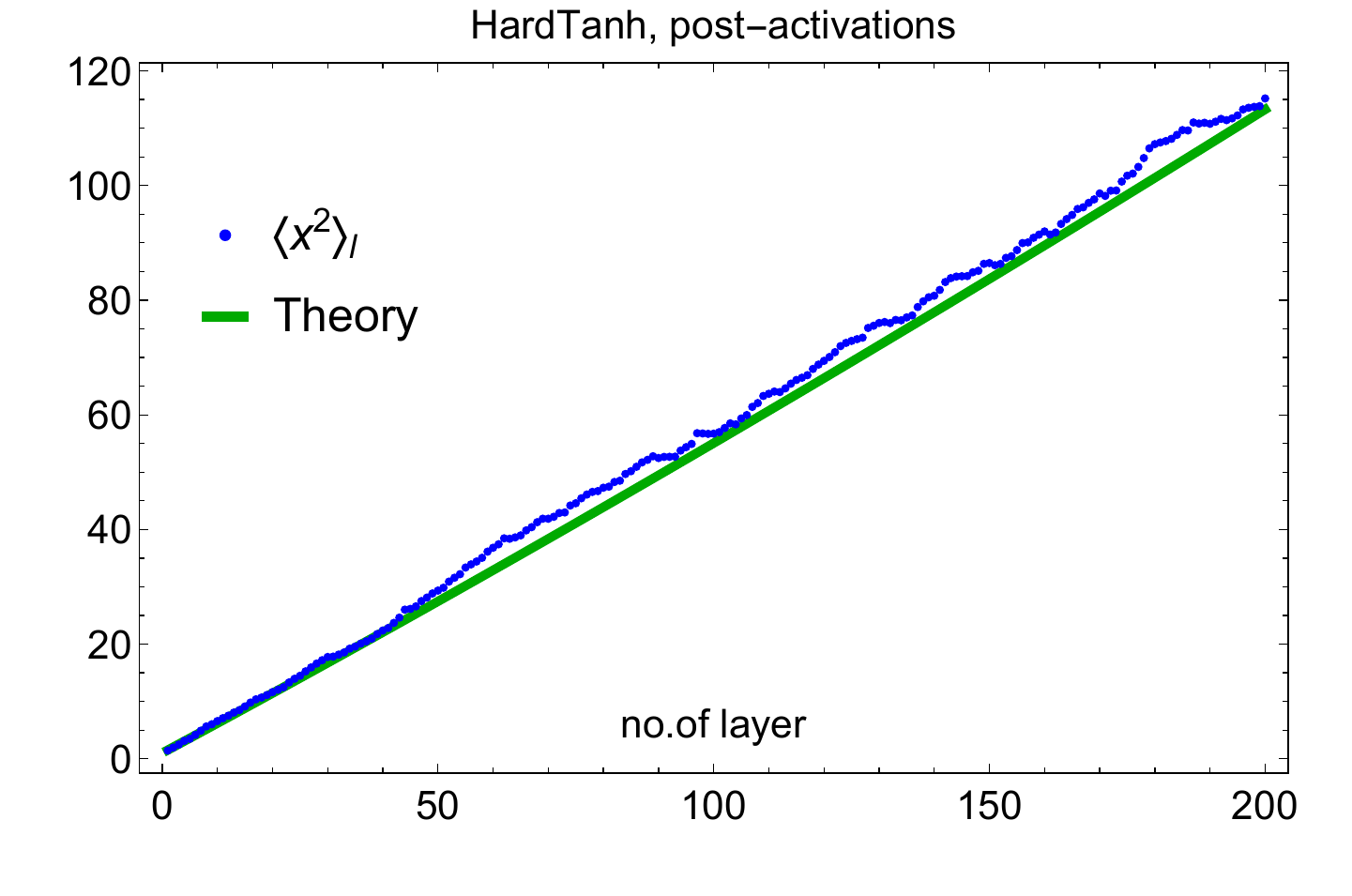}\includegraphics[width=0.33\textwidth]{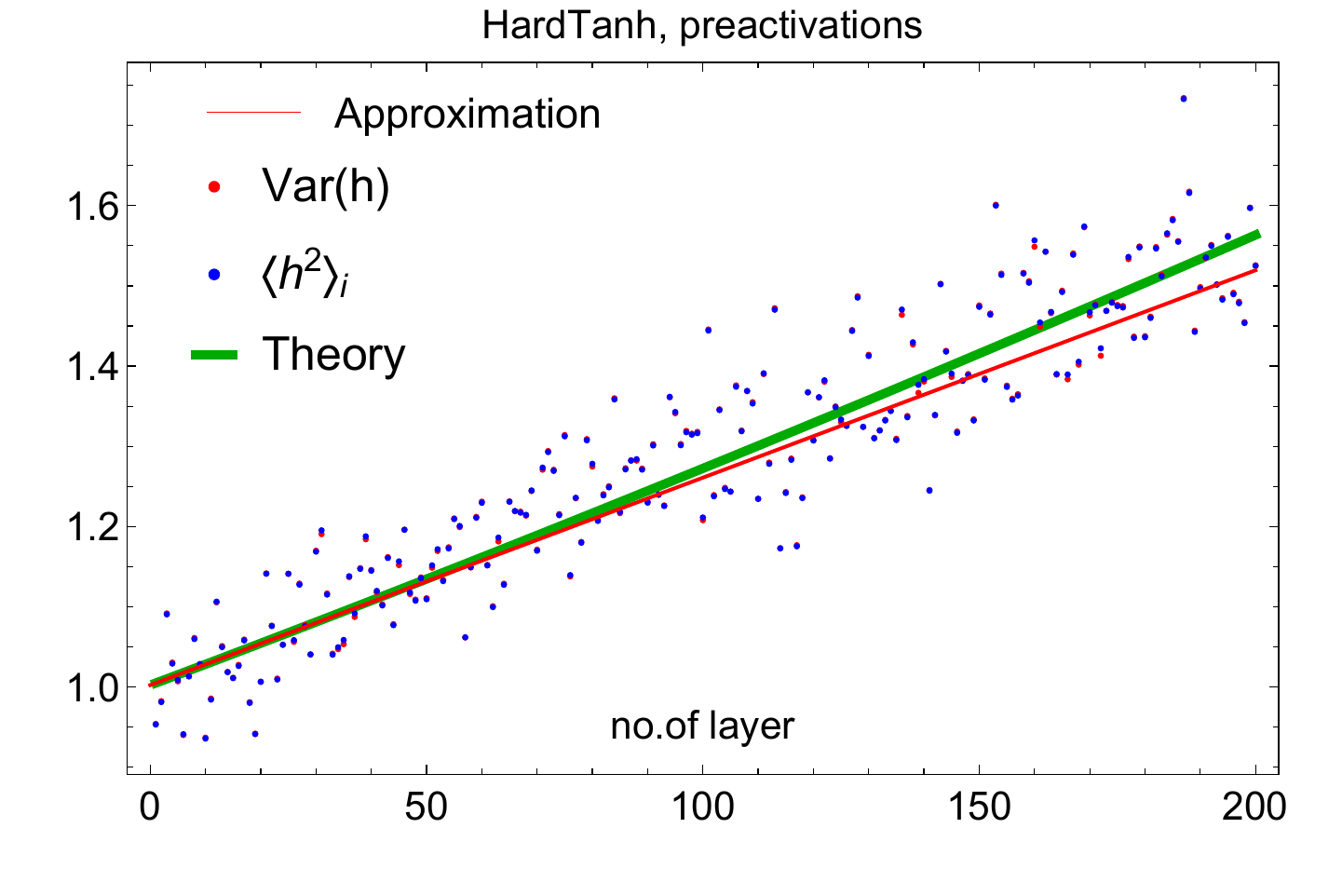}\includegraphics[width=0.33\textwidth]{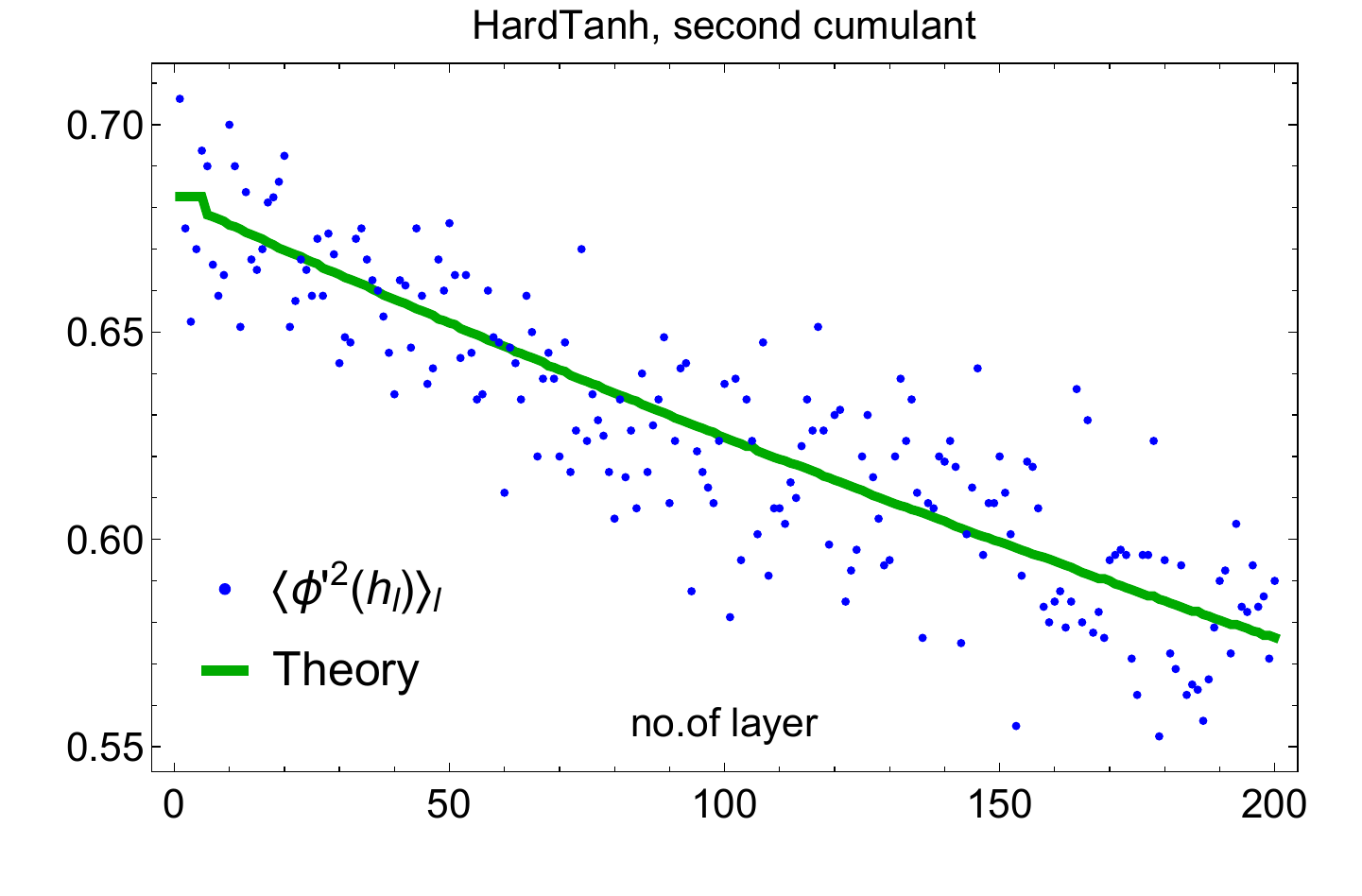}
\includegraphics[width=0.33\textwidth]{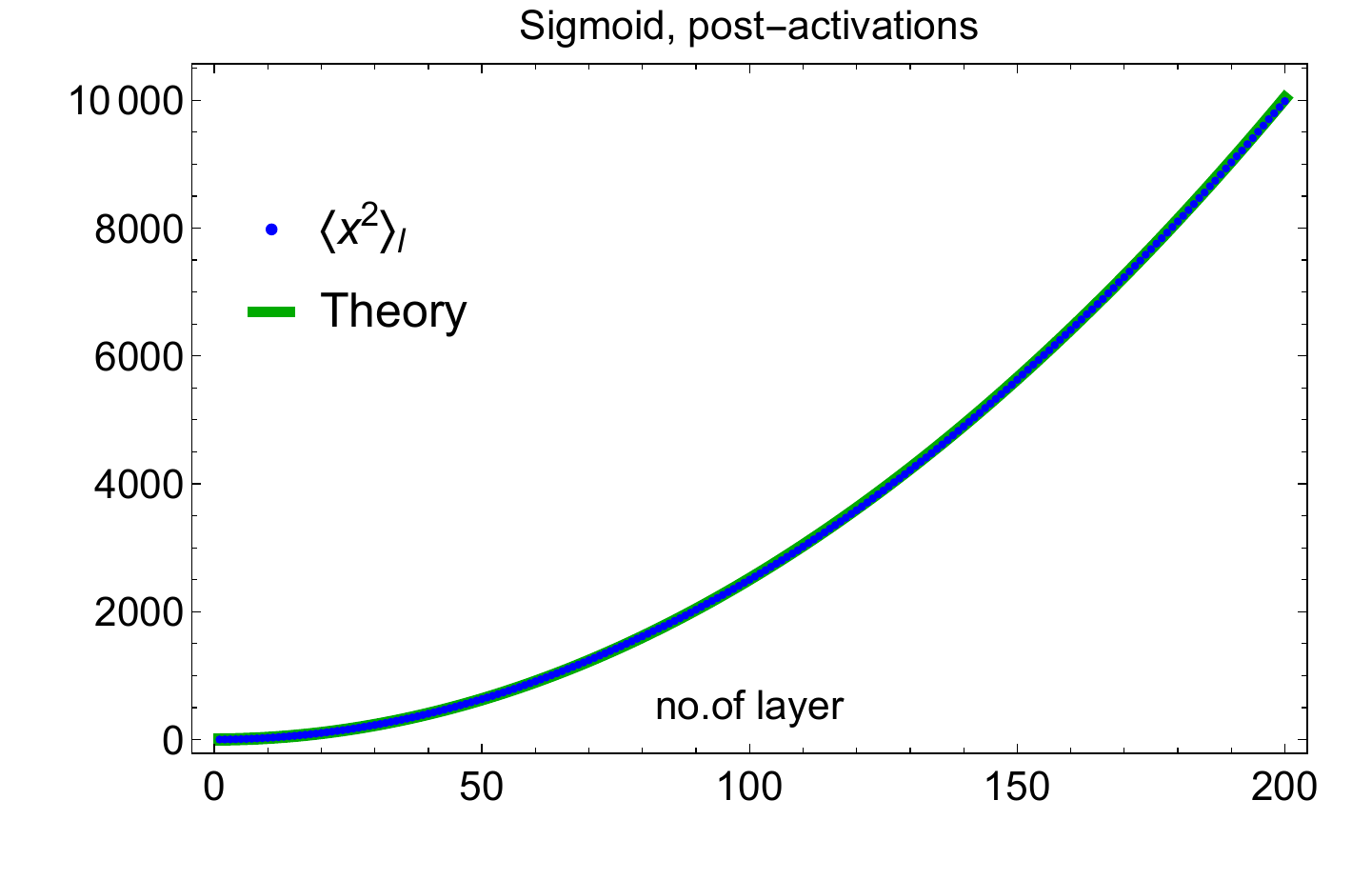}\includegraphics[width=0.33\textwidth]{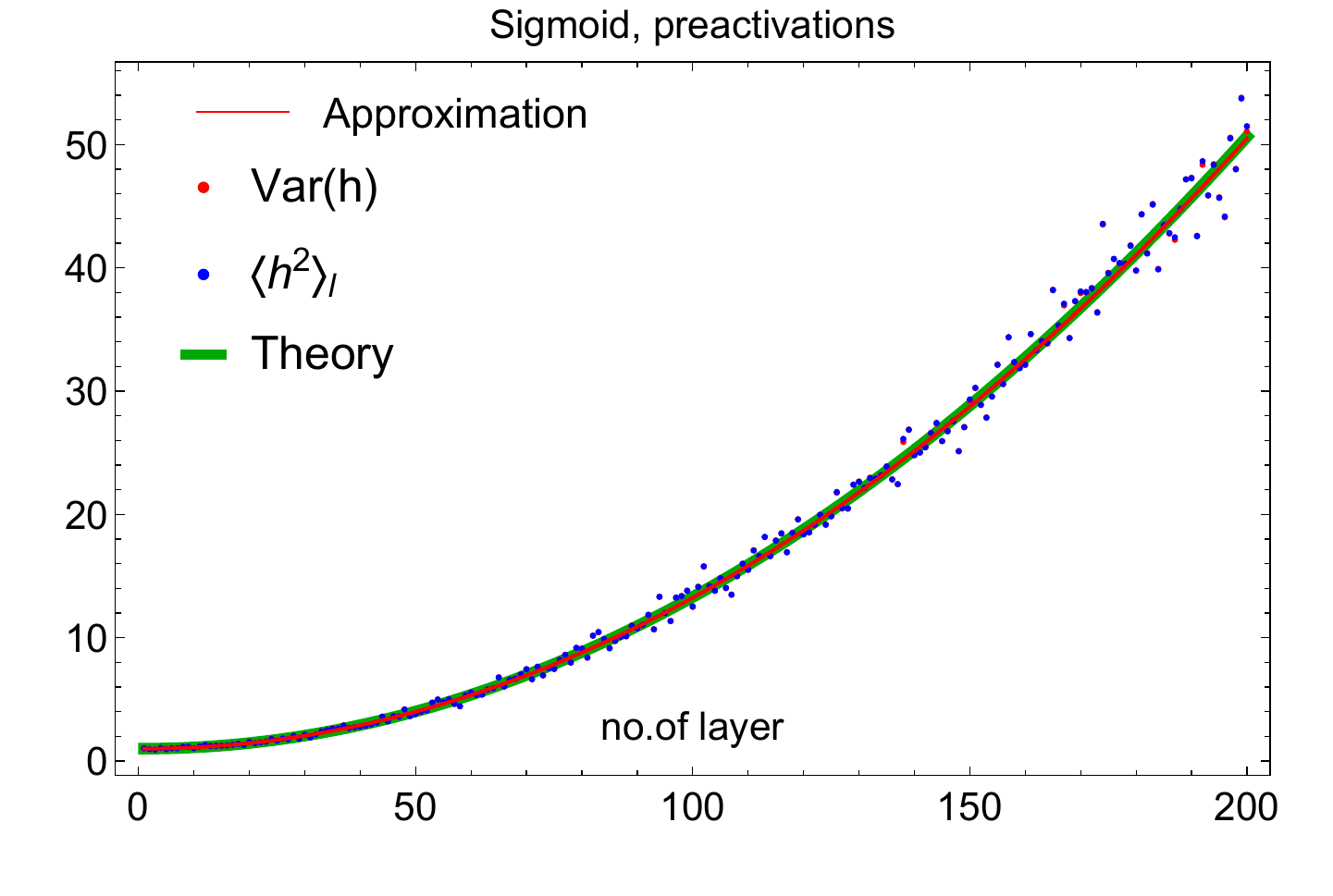}\includegraphics[width=0.33\textwidth]{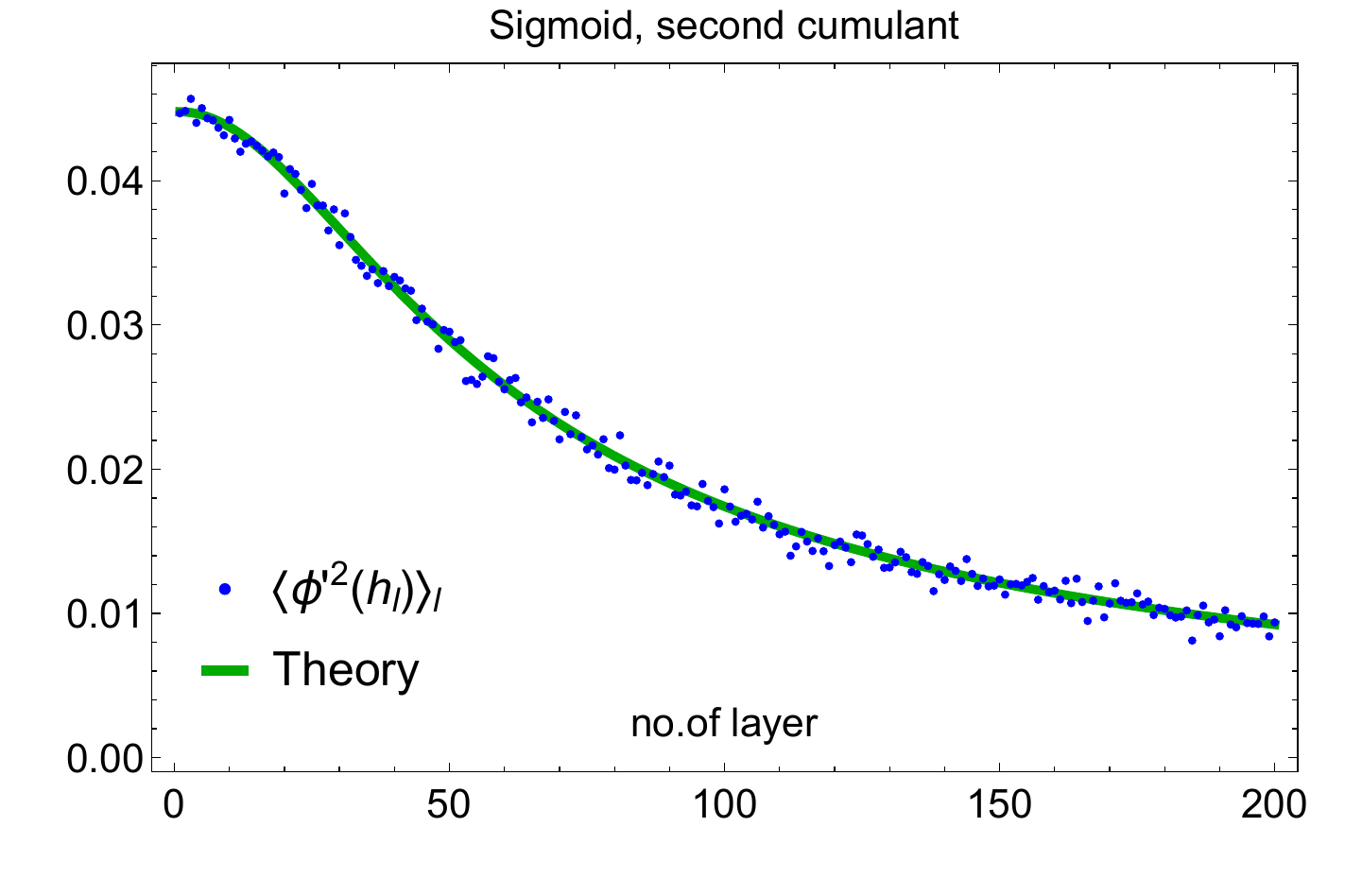}
\\
\includegraphics[width=0.33\textwidth]{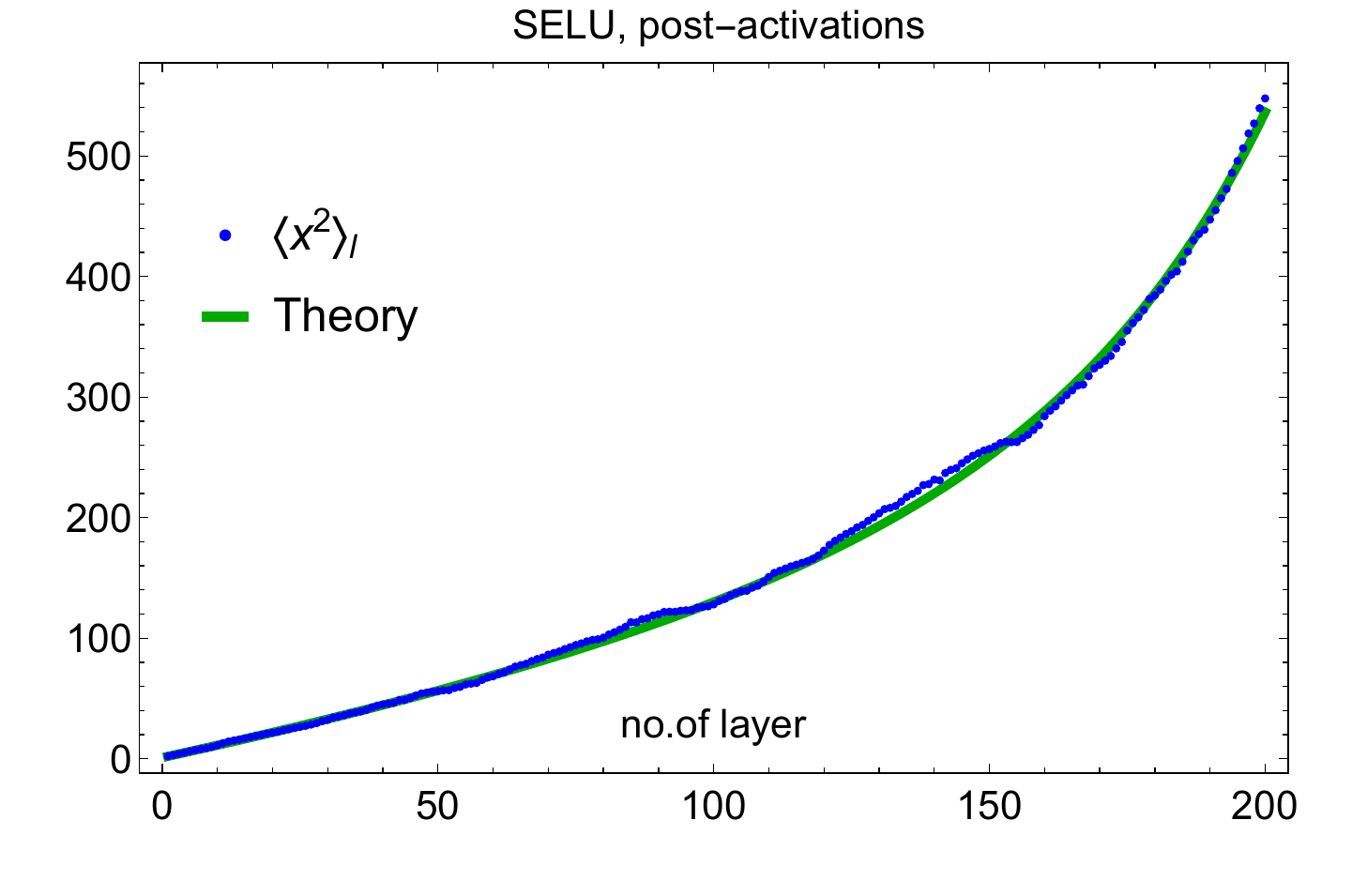}\includegraphics[width=0.33\textwidth]{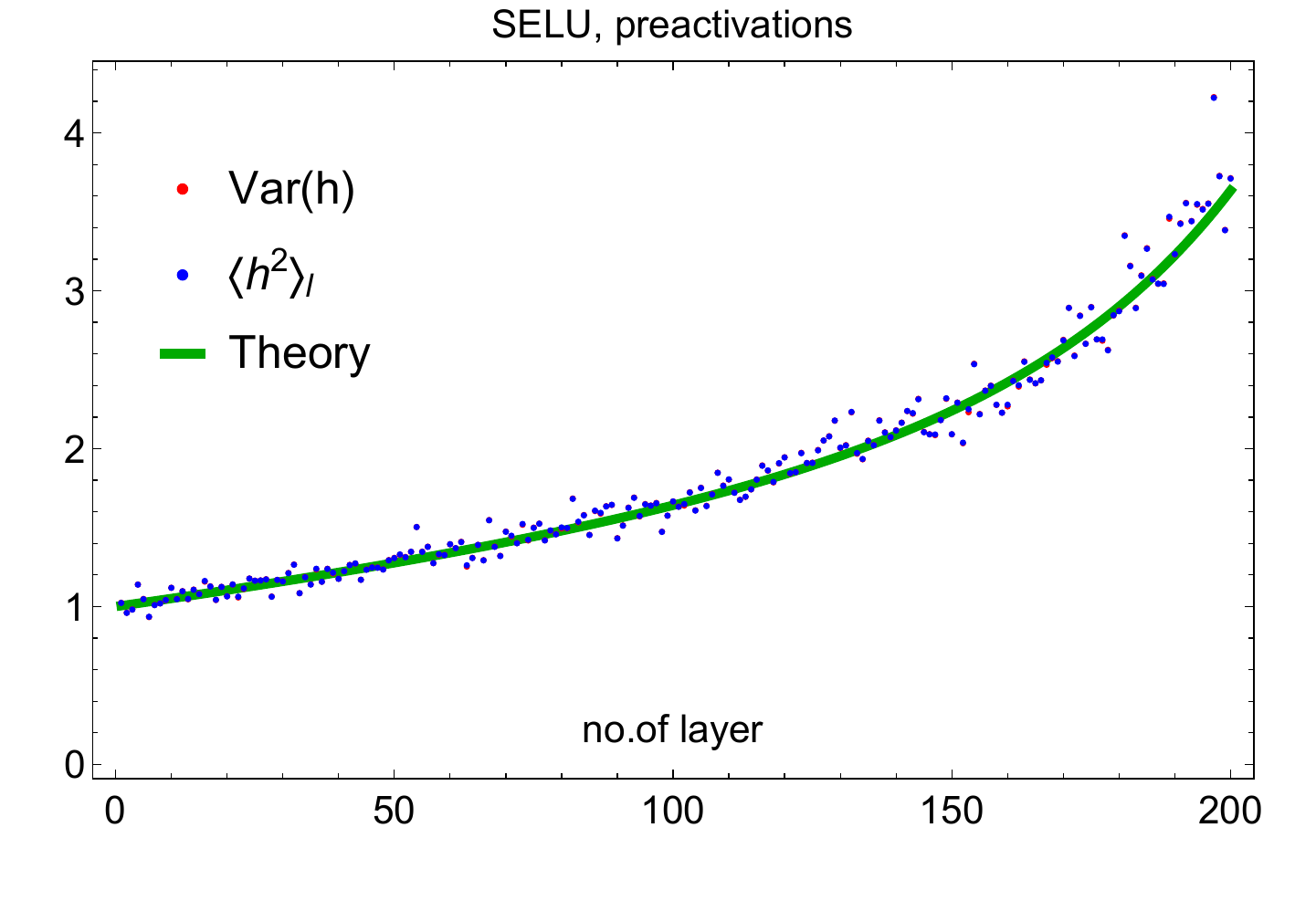}\includegraphics[width=0.33\textwidth]{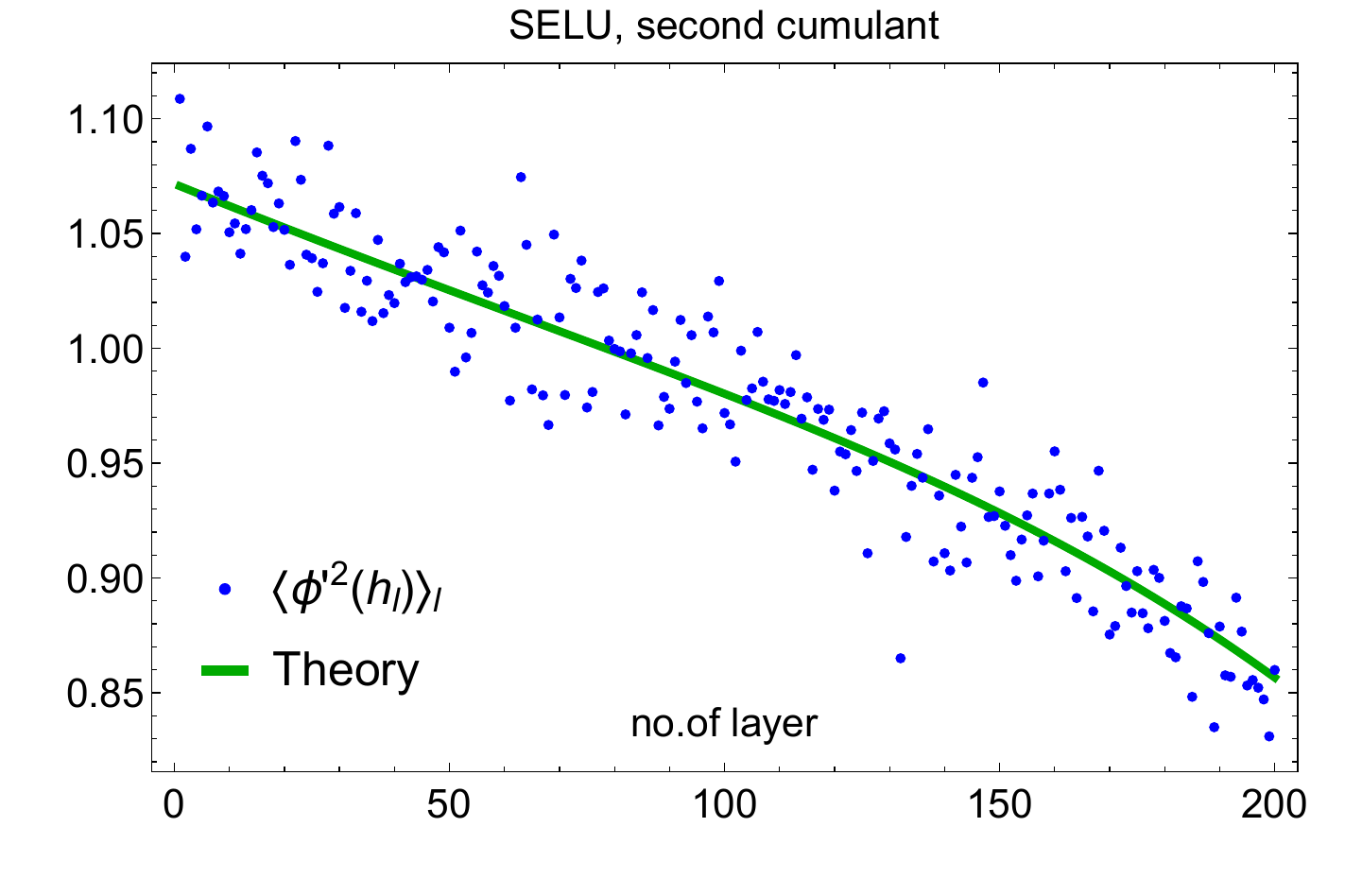}

\caption{Verification of the numerical solution to the recurrence equations for post-activations \eqref{eq:x2_1} (left column) and preactivations \eqref{eq:recursion} (middle). Based on this signal propagation the effective cumulant $c_l$ for each layer was calculated (right column).
The solid red lines represent the approximation \eqref{eq:Approx} for tanh and hard tanh nonlinearity and \eqref{eq:recursion_sigmoid_approx} for sigmoid. Solutions of recurrences (solid) are confronted with the numerical simulation (dots) of residual fully connected networks with $L=200$ layers of width $N=800$. Data points represent a single run of simulations. Weights are independently sampled from Gaussian distribution of zero mean and variance equal to $\frac{1}{NL}$. Biases and network input are sampled from standard normal distribution. A small variability of $c^l$ across the network justifies the assumption made for derivation of~\eqref{eq:Sapprox}.
\label{Fig:Verification2}
}
\end{figure}

\begin{figure}
\includegraphics[width=0.33\textwidth]{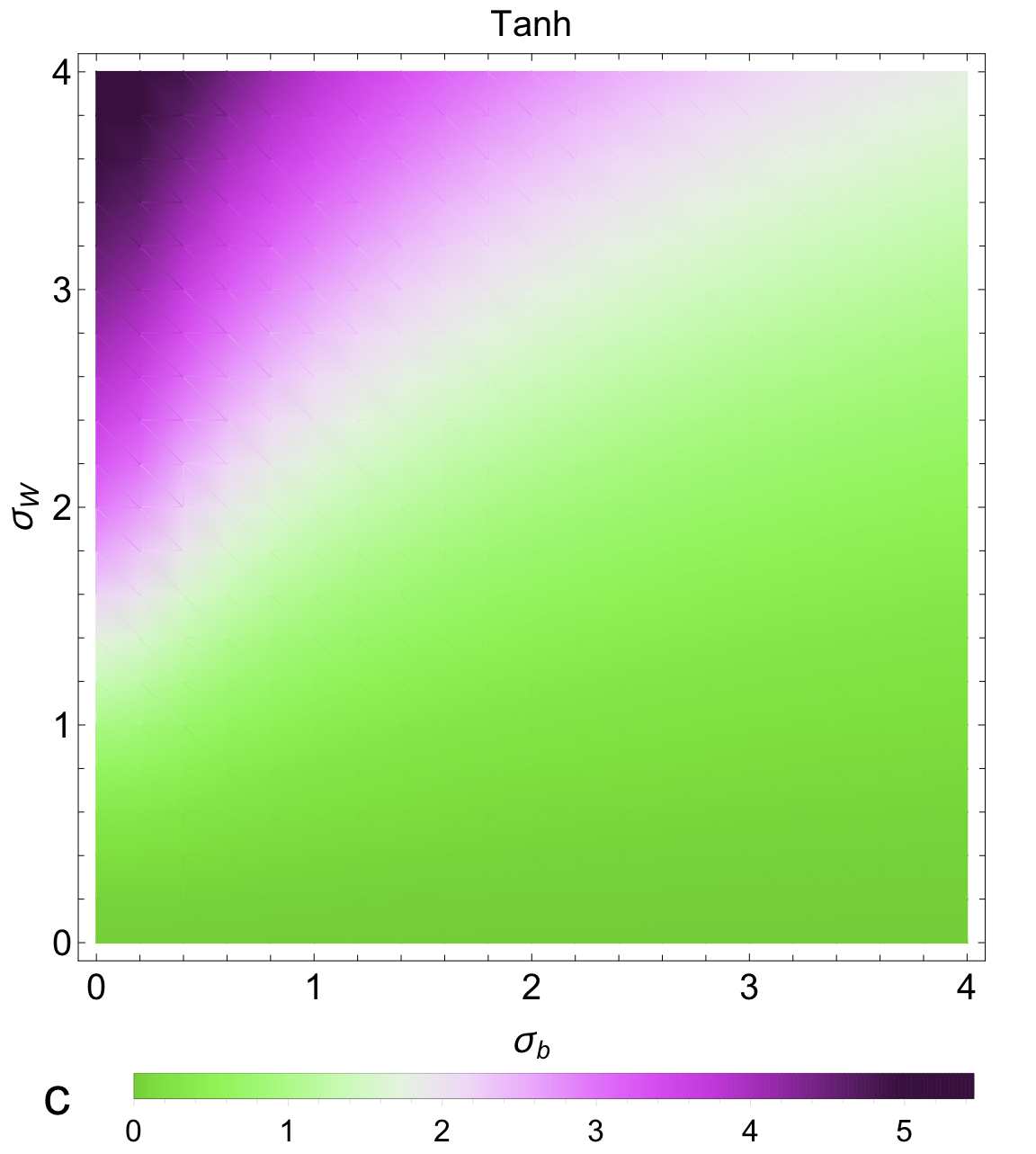}\includegraphics[width=0.33\textwidth]{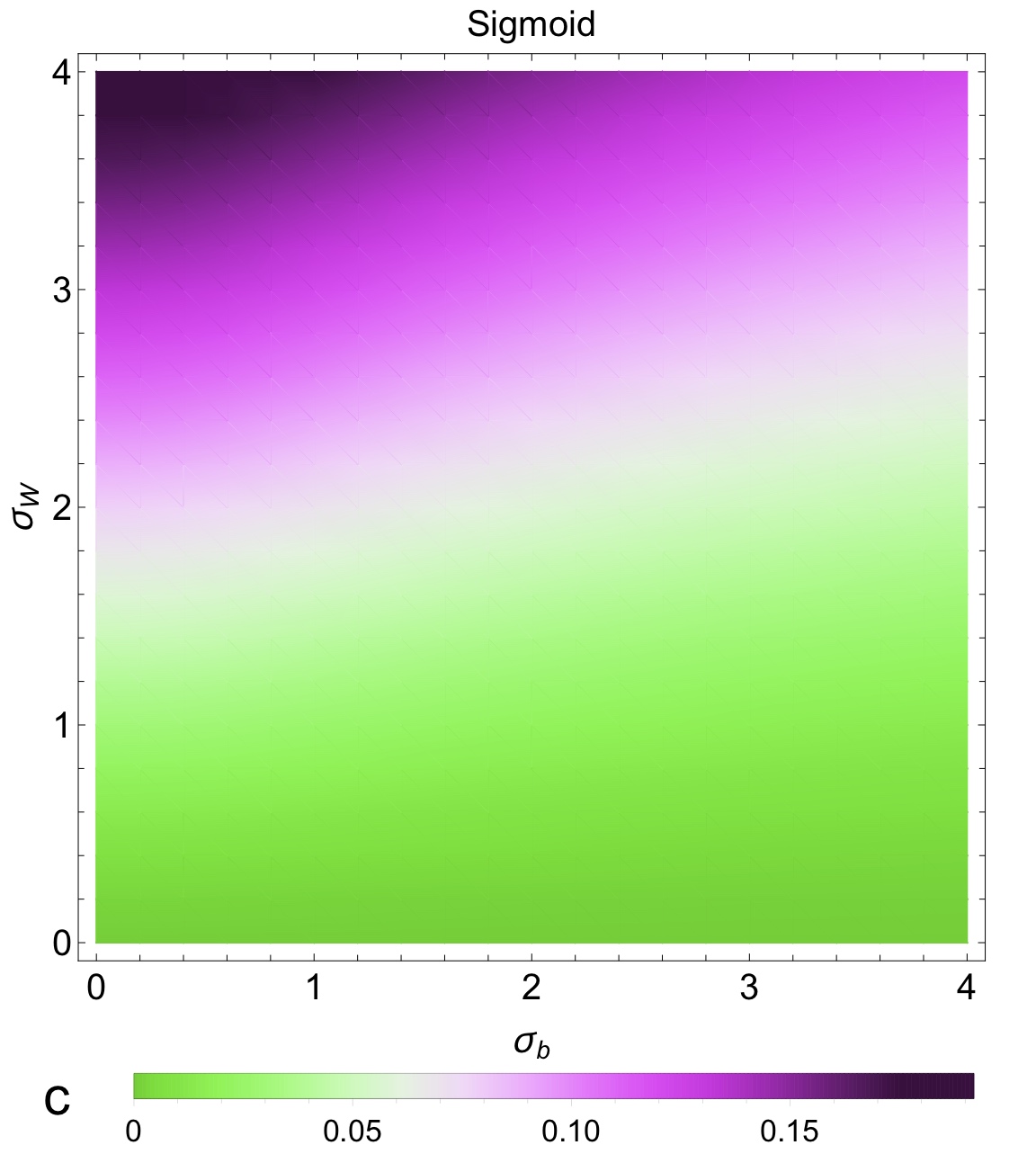}\includegraphics[width=0.33\textwidth]{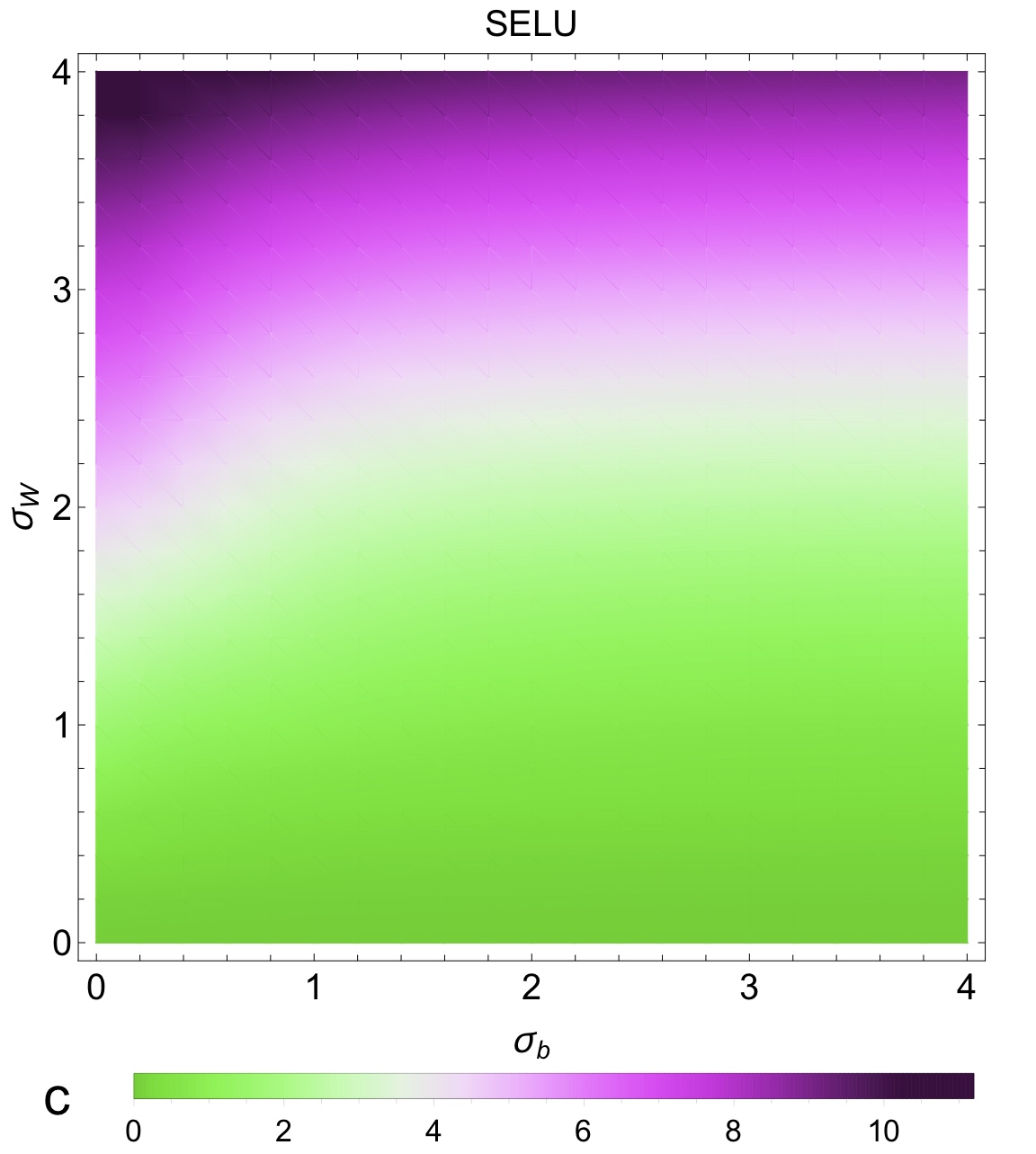}
\caption{Dependence of the parameter $c$ (which determines the shape of the spectrum) on the variances of biases and weights. The smaller $c$, the closer to the perfect dynamical isometry. Note different scales on each plot. Low value of $c$ for sigmoid is a consequence of saturation of the nonlinearity.\label{Fig:Cparam}}
\end{figure}

To validate the assumptions made and corroborate the theoretical results obtained  in subsections \ref{s:signal} and \ref{s:actfunc}, we simulate signal propagation in the studied residual neural networks with different activation functions. The outcomes of these experiments are depicted in Figure \ref{Fig:Verification2}. Numerical solution to the recurrence relations allows us to numerically calculate the parameter $c$, as a function of variances of weights and biases, which is presented in Figure~\ref{Fig:Cparam}. 

\section{Baseline}
\label{a:2}

\begin{figure}[ht]\begin{center}
\includegraphics[width=0.556\textwidth]{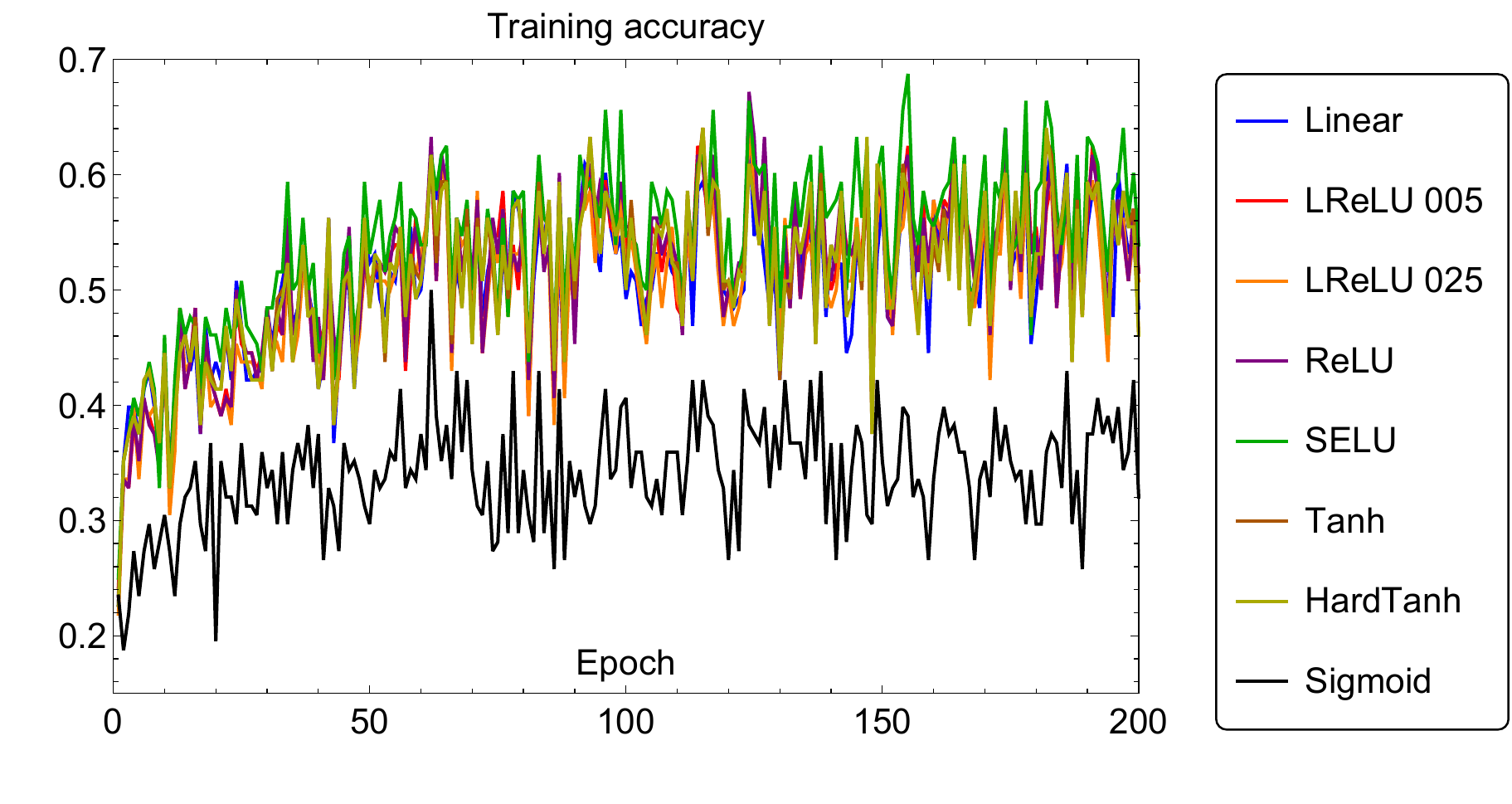}\includegraphics[width=0.444\textwidth]{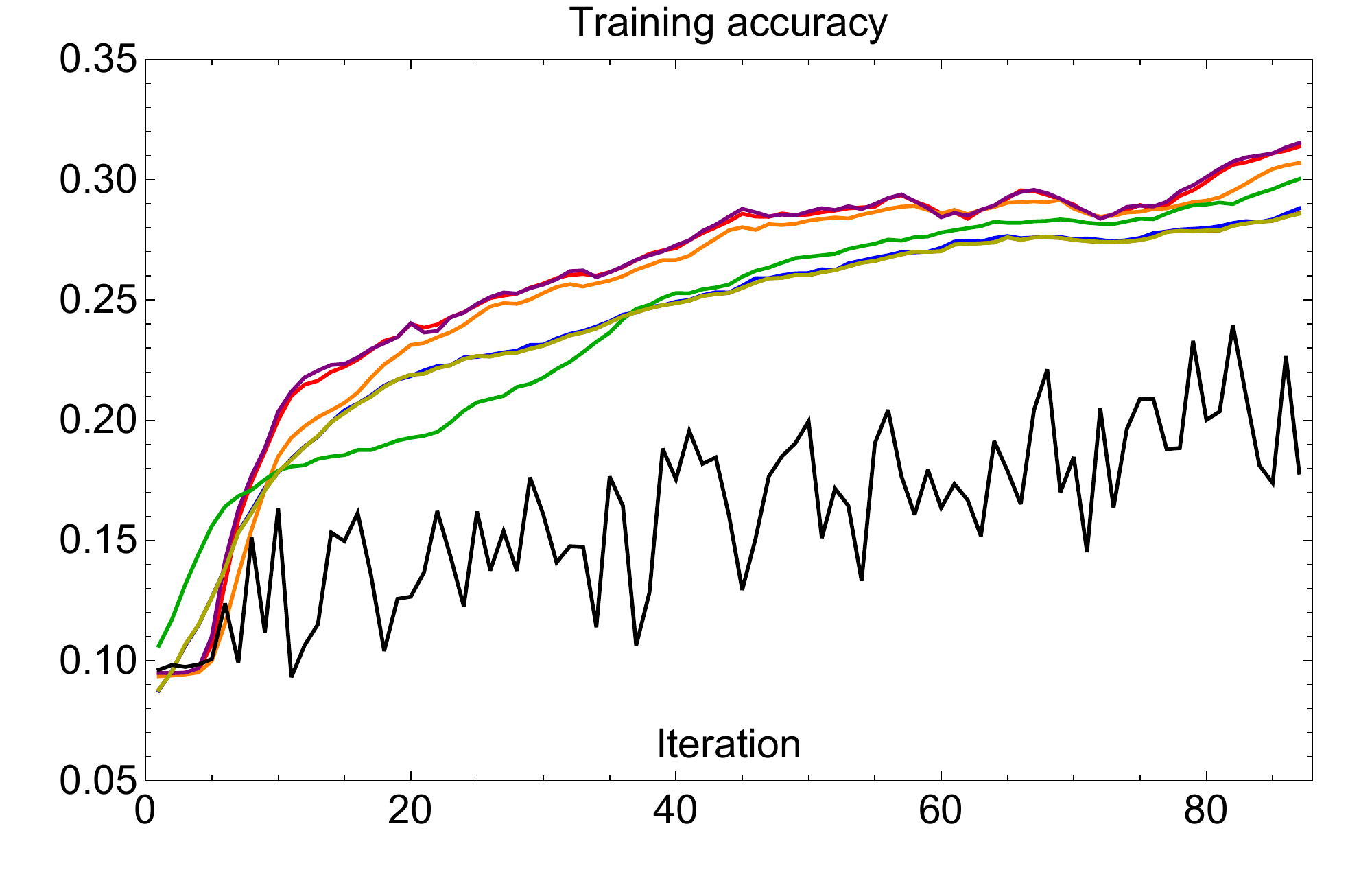} \\
\includegraphics[width=0.49\textwidth]{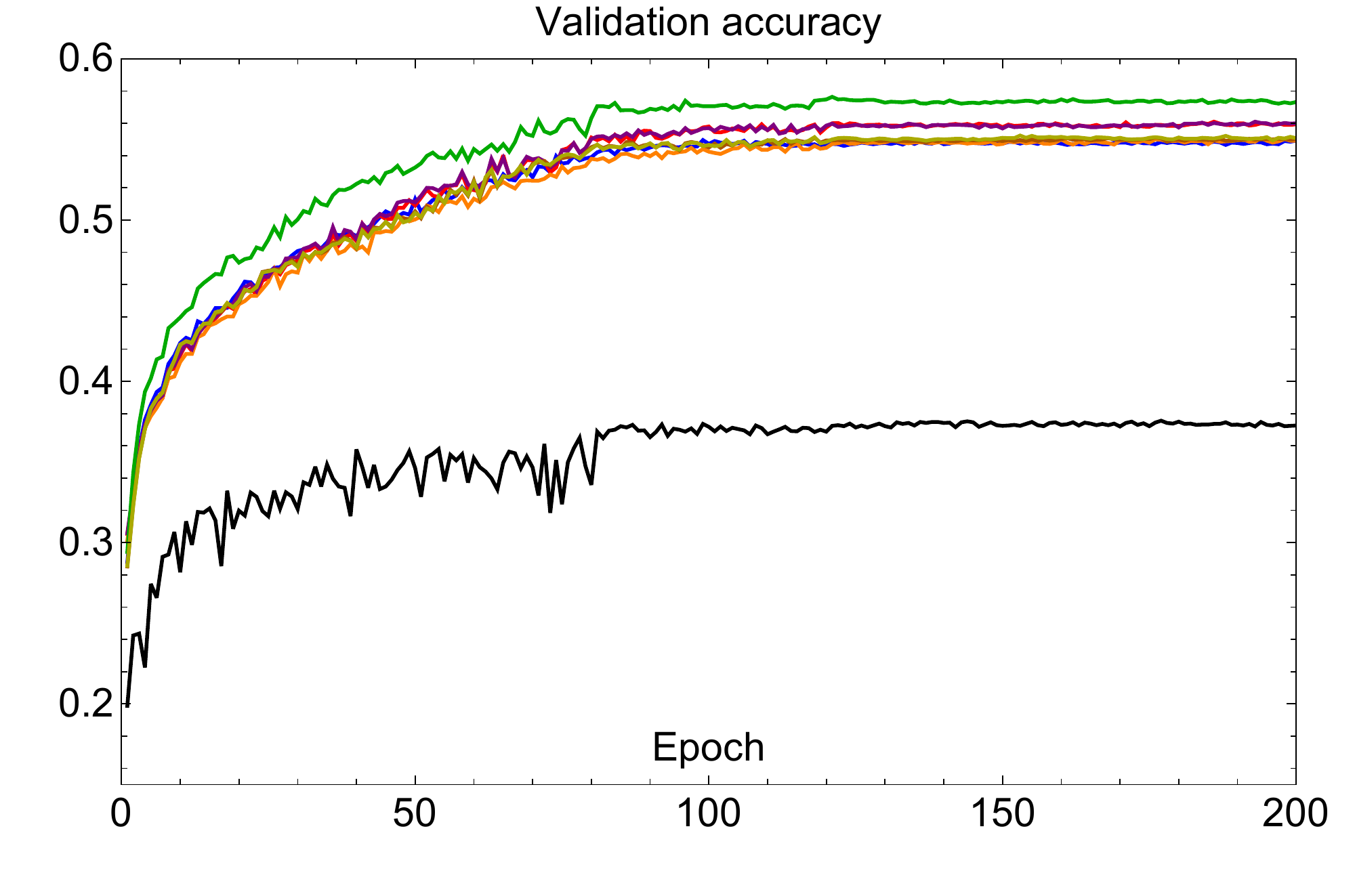}\includegraphics[width=0.49\textwidth]{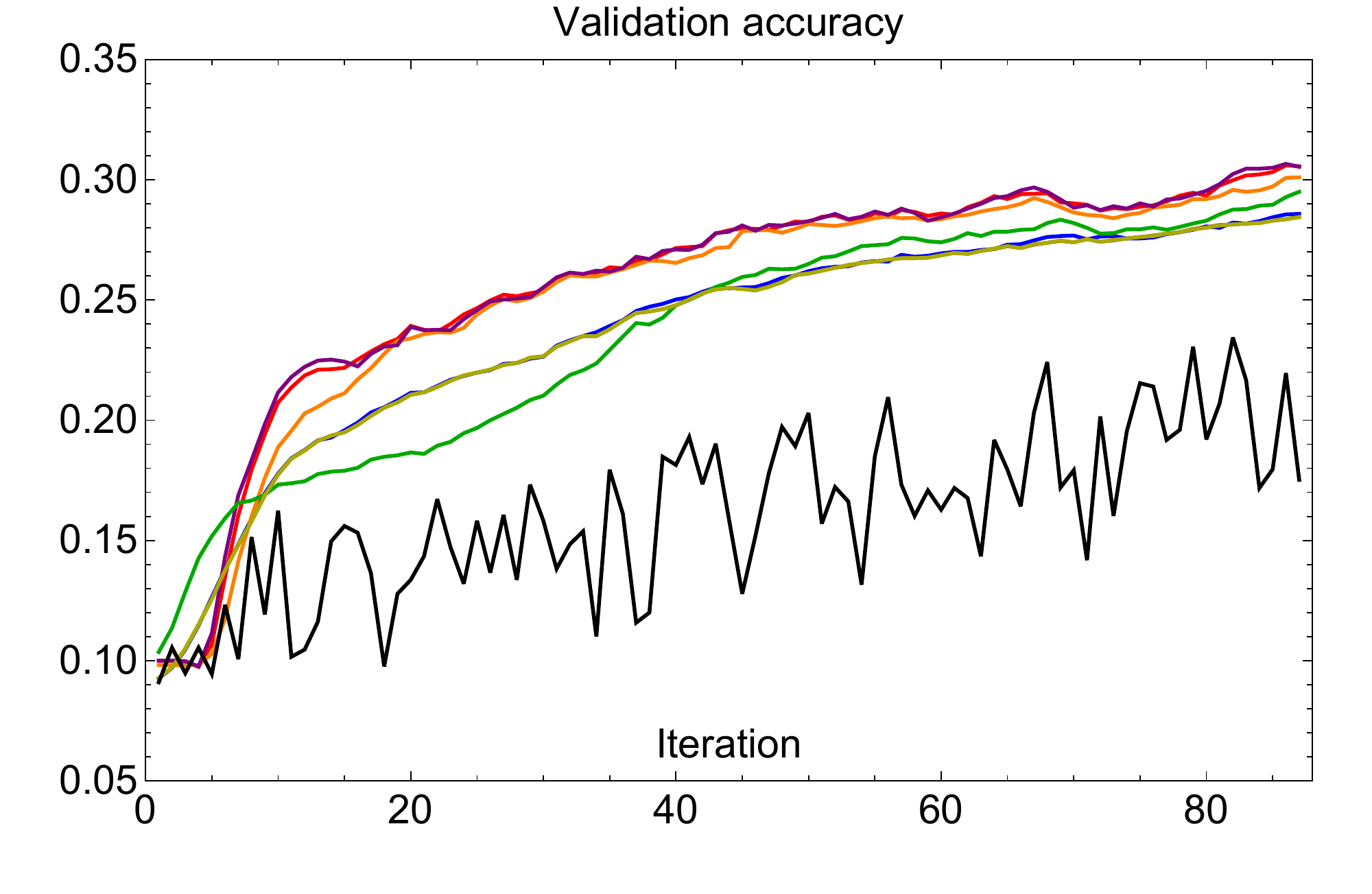}
\end{center}
\caption{Training (top) and validation (bottom) accuracy during first 200 epochs (left) and first 100 iterations (right) of residual networks with various activation functions. The weight initialization was Gaussian with zero mean and $1/NL$ variance. We set $\alpha=0.05$ and $\alpha=0.25$ for leaky ReLU (LReLU). \label{Fig:baseline}  }
\end{figure}

We advocate for setting the same value of the effective cumulant (and hence keeping the same spectrum of the input-output Jacobian)
when comparing the effects of using different activation function on the learning process. Thus, here, in Fig. \ref{Fig:baseline}, for comparison, we showcase the learning accuracy when instead of the effective cumulant, the weight matrices entries' variances (equal to $1/NL$) are kept the same across the networks.

\end{document}